%% file: main.tex
\documentclass{article}

\usepackage{arxiv}

\usepackage{amsmath}
\usepackage[utf8]{inputenc} %
\usepackage[T1]{fontenc}    %
\usepackage{hyperref}       %
\usepackage{url}            %
\usepackage{booktabs}       %
\usepackage{amsfonts}       %
\usepackage{nicefrac}       %
\usepackage{microtype}      %
\usepackage{cleveref}       %
\usepackage{lipsum}         %
\usepackage{graphicx}
\usepackage{float}
\usepackage{natbib}
\setcitestyle{authoryear, square}
\setcitestyle{authoryear}
\bibpunct{[}{]}{;}{a}{}{,}
\renewcommand{\cite}[1]{\citep{#1}}

\usepackage{doi}
\usepackage{xcolor}
\usepackage{tikz}
\usepackage[edges]{forest}
\definecolor{hidden-draw}{RGB}{205, 44, 36}
\definecolor{hidden-blue}{RGB}{194,232,247}
\definecolor{hidden-orange}{RGB}{243,202,120}
\definecolor{hidden-yellow}{RGB}{242,244,193}
\definecolor{tree-level-1}{RGB}{245,20,85}
\definecolor{tree-level-2}{RGB}{246,86,118}
\definecolor{tree-level-3}{RGB}{248,177,193}
\definecolor{tree-leaf}{RGB}{176,230,198}

\definecolor{Self}{RGB}{255,0,128}
\definecolor{Ensemble}{RGB}{0,127,255}
\definecolor{Iterative}{RGB}{153,51,255}

\definecolor{exemplar1}{RGB}{136,98,148}
\definecolor{exemplar2}{RGB}{148,210,242}
\definecolor{knowledge1}{RGB}{249,219,152}
\definecolor{knowledge2}{RGB}{255,245,220}

\usepackage{subcaption}
\usepackage{pgfplots}
\pgfplotsset{compat=1.16}
\usetikzlibrary{pgfplots.groupplots}

\input{packages_and_commands.tex}

\title{\input{LaTeX/title.tex}}

\date{}
\author{Shreyas Chaudhari\textsuperscript{*1} \And Pranjal Aggarwal\textsuperscript{*2} \And Vishvak Murahari\textsuperscript{3} \And Tanmay Rajpurohit\textsuperscript{4} \And Ashwin Kalyan\textsuperscript{5}  \quad  Karthik Narasimhan\textsuperscript{3} \quad Ameet Deshpande\textsuperscript{3} \quad Bruno Castro da Silva\textsuperscript{1} \\ \\
        \textsuperscript{1}University of Massachusetts Amherst \\
        \textsuperscript{2}Department of Computer Science, Indian Institute of Technology, Delhi \\
        \textsuperscript{3}Department of Computer Science, Princeton University \\
        \textsuperscript{4}Georgia Tech \\
        \textsuperscript{5}Independent Researcher \\
        \textsuperscript{*}Equal Contribution \\
        \texttt{schaudhari@cs.umass.edu, pranjal2041@gmail.com}}

\renewcommand{\headeright}{}
\renewcommand{\undertitle}{}
\renewcommand{\shorttitle}{}

\allowdisplaybreaks
\begin{document}
\maketitle

\input{LaTeX/000_abstract.tex}

\input{LaTeX/100_introduction.tex}

\input{LaTeX/analysis.tex}

\input{LaTeX/survey.tex}

\input{LaTeX/conclusion.tex}

\bibliographystyle{plainnat}
\bibliography{survey}

\end{document}

%% file: packages_and_commands.tex
\definecolor{mydarkblue}{rgb}{0,0.08,0.45}

\usepackage{hyperref}
\hypersetup{
    colorlinks=true,
    linkcolor=blue,
    filecolor=magenta,
    urlcolor=cyan,
    citecolor=mydarkblue,
}
\usepackage[normalem]{ulem}

\newcommand{\commentblock}[1]{}

\newcommand{\shreyas}[1]{\textcolor{teal}{[Shreyas: #1]}}

\newcommand{\future}[1]{\textcolor{green}{}}

\newcommand{\cmmnt}[1]{\ignorespaces}

\newcommand{\needcite}{\textcolor{blue}{[Link Citation]} }

\newcommand{\pipre}{\pi_\text{pre}}
\newcommand{\pirlhf}{\pi_\text{rlhf}}
\newcommand{\E}{\mathbb{E}}
\newcommand{\Ppre}{P_\text{pre-train}}

%% file: LaTeX/title.tex
RLHF Deciphered: \\A Critical Analysis of Reinforcement Learning from Human Feedback for LLMs

%% file: LaTeX/000_abstract.tex
\begin{abstract}
    State-of-the-art large language models (LLMs) have become indispensable tools for various tasks.
    However, training LLMs to serve as effective assistants for humans requires careful consideration.
    A promising approach is reinforcement learning from human feedback (RLHF), which leverages human feedback to update the model in accordance with human preferences and
    mitigate issues like toxicity and hallucinations.
    Yet, an understanding of RLHF for LLMs is largely entangled with initial design choices that popularized the method and current research focuses on augmenting those choices rather than fundamentally improving the framework.
    In this paper, we analyze RLHF through the lens of reinforcement learning principles to develop an understanding of its fundamentals,
    dedicating substantial focus to the core component of RLHF---the reward model.
    Our study {investigates} modeling choices, caveats of function approximation, and their implications on RLHF training algorithms, highlighting the underlying assumptions made about the expressivity of reward.
    Our analysis improves the understanding of the role of reward models and methods for their training, concurrently revealing limitations of the current methodology.
    We characterize these limitations, including incorrect generalization, model misspecification, and the sparsity of feedback, along with their impact on the performance of a language model.
    The discussion and analysis are substantiated by a categorical review of current literature, serving as a reference for researchers and practitioners to understand the challenges of RLHF and build upon existing efforts.
\end{abstract}

%% file: LaTeX/100_introduction.tex
\newpage

{\hypersetup{linkcolor=black}\tableofcontents}

\newpage

\section{Introduction}
\label{sec:introduction}

\begin{figure}[t]
    \centering
    \includegraphics[width=0.95\textwidth]{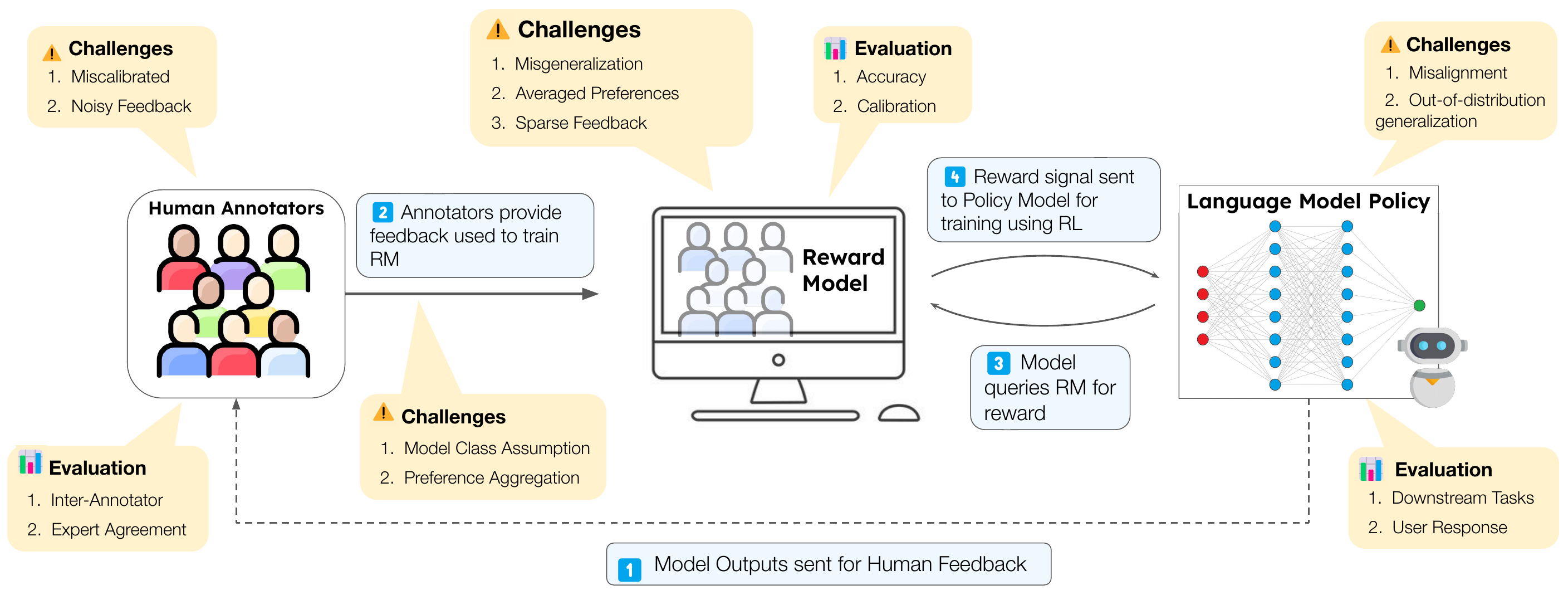}
    \caption{Overview of the RLHF procedure, illustrating the challenges encountered at each step.
        The paper conducts a detailed examination of these challenges, providing valuable insights into each stage of the procedure.
    }
    \label{fig:paper_figure}
\end{figure}

Large Language Models (LLMs) demonstrate remarkable capabilities that extend beyond basic language tasks, leading to their widespread adoption across various industries.
The remarkable utility of these models holds the potential to transform established workflows in critical sectors such as technology, healthcare, finance, and education~\cite{Singhal2022LargeLM, Wu2023BloombergGPTAL, Yan_2023}.
As they become integral to these domains, it's crucial to ensure that the behavior of LLMs is predictable, safe, and trustworthy—--meeting the expectations set for a human performing the same tasks.
This challenge of making LLMs exhibit human-like qualities, known as \textit{alignment} with human objectives, is central to making these models suitable for diverse tasks.
An effective method for addressing this challenge is reinforcement learning from human feedback (RLHF).

RLHF first gained popularity due to its ability to solve reinforcement learning (RL) problems like simulated robotic locomotion and playing Atari games \cite{DeepRLHF-Christiano2017DeepRL} without access to a reward function, by simply leveraging human feedback about preferences on demonstrated behaviors.
It has since been adopted for fine-tuning LLMs using human feedback.
This leads to a natural inquiry: {How can a method designed to master games be effectively used to align LLMs with human objectives?}
The method has proven to be immensely successful \cite{OpenAI2022Chatgpt}, but not without well-documented limitations \cite{casper2023open}.
A comprehensive understanding of \textit{why} it achieves its success remains largely elusive.
Consequently, research efforts on the topic are stuck in a local minima, with variants focused on augmenting the components of the method---including the training algorithm \cite{Ramamurthy2022IsRL}, reward model \cite{FineGrainedRLHF-Wu2023FineGrainedHF}, and even RL-free approaches \cite{rafailov2023direct}.
However, some fundamental limitations of the approach remain obscured due to the overarching goal of recent work to refine the initial design choices.

In this work, we develop a comprehensive understanding of RLHF by analyzing the core components of the method.
We begin the study by motivating the necessity for RLHF by highlighting the problem of objective mismatch in pre-trained LMs (Section \ref{sec:motivation}).
To formulate foundational questions about the framework, we adopt a Bayesian perspective of RLHF.
It serves to highlight the significance of the reward function in particular (Section \ref{sec:role_of_reward}).
The reward function forms the central cog of the RLHF procedure, and the design choices used to model it form a major focus of our study.

\commentblock{The substantial advancements in LLM capabilities, predominantly achieved through extensive scaling and generative pre-training~\cite{Brown2020LanguageMA, Anil2023PaLM2T}, significantly shape their demonstrated capabilities.
    Fundamentally, pre-trained language models suffer from an \textit{objective mismatch} problem.
    The pre-training objective often deviates from the specific requirements of tasks to which they are applied, especially when capabilities beyond language modeling are necessary for solving the task (Section \ref{sec:motivation}).
    Consequently, methods such as RLHF~\cite{InstructGPT-Ouyang2022TrainingLM} become essential for fine-tuning the parameters of the language model, enabling the generation of outputs aligned with human objectives for addressing a task.

    Recent advancements aimed at enhancing the effectiveness of RLHF concentrate on refining the initial design choices that popularized the method \cite{Ramamurthy2022IsRL}.
    \needcite
    The resultant effect has been an entanglement of the understanding of RLHF with those specific design choices, encompassing the training algorithms, reward modeling paradigms, and more.
    This inadvertently overlooks essential aspects of the procedure such as the characteristics of the reward model (Section \ref{sec:rlhf_reward_function}).
    The reward model stands at the core of the functioning of RLHF.
    Within the realm of reinforcement learning, extensive research delves into reward shaping and reward inference, which is now beginning to be leveraged for LLMs.
}

The current formulation of RLHF relies on a set of assumptions to model the reward function (Section \ref{sec:oracular_reward}, \ref{sec:btl_intro}).
Following the delineation of these assumptions, an analysis of the reward model independent of specific modeling choices follows.
The analysis, in a principled manner, provides an understanding of issues such as:
\begin{enumerate}
    \item The impractical requirement for extensive amounts of feedback data for training accurate reward models.
    \item The combination of very limited feedback data and the use of function approximation results in misgeneralization, wherein inaccurate reward values are assigned to inputs not seen during training.
\end{enumerate}
These imperfections of the reward model, along with challenges such as reward sparsity and reward model misspecification, are highlighted in the paper (Section \ref{sec:analysis_reward_model}).
Their impact on the performance of a language model is explored in detail (Section \ref{sec:misalignment}).
The course of the analysis leads to the formalization of concepts such as an \textit{oracular reward} that serve as the theoretical golden standard for future efforts (Section \ref{sec:oracular_reward}).
An overview of the RLHF procedure along with the various challenges studied in this work is provided in Figure \ref{fig:paper_figure}.

The discussion is followed by an extensive survey of an expanding body of literature related to the topic.
The survey is organized into sections that outline the framework of RLHF.
Starting with a high-level overview of Large Language Models (LLMs), the survey systematically covers various aspects:
\begin{itemize}
    \item Different types of human (and non-human) feedback (Section \ref{sec:human_feedback}),
    \item The training methods in RLHF (Section \ref{sec:rlhf_training}),
    \item Alternative approaches that do not rely on RL or reward models (Section \ref{sec:alternatives}).
\end{itemize}
This structure aims to provide a comprehensive overview of the extensive landscape of works that have contributed to the remarkable success of RLHF.

\commentblock{%
    As the attainment of superintelligence~\cite{leike2023IntSuper, Bostrom_2017} looms closer
    and research on superalignment becomes imperative, comprehending the limits of RLHF in its current state is crucial.
    Our insights aim to expand the boundary of the utility of human supervision and provide a conceptual framework for measuring the degree of alignment.

    The discussion begins by highlighting the issue of objective mismatch in pre-trained language models (Section \ref{sec:motivation}).
    This is followed by a formulation of the LM as an agent acting in a Markov decision process (Section \ref{sec:llm_rl_formulation}) serving to highlight the role of the reward.
    A gold standard reward is first conceptualized (Section \ref{sec:oracular_reward}), and then existing practical methods for estimating rewards and their practical limitations are studied (Section \ref{sec:rlhf_reward_function}).
    This serves to highlight the drawbacks of existing approaches and identify avenues for future improvements.
    The discussion culminates in highlighting the effectiveness of reinforcement learning as a means to achieve alignment.
    This is followed by an extensive review of recent literature on the topic, along with pivotal older literature from areas like reinforcement learning that tackle analogous problems, providing additional insight into the problem of alignment in LLMs.}

%% file: LaTeX/analysis.tex
\input{LaTeX/analysis_sections/2-motivation}

\input{LaTeX/analysis_sections/3-rl-formulation}

\input{LaTeX/analysis_sections/4-reward-learning}

\input{LaTeX/analysis_sections/5-estimated-reward}

\input{LaTeX/analysis_sections/6-rl-on-reward}

%% file: LaTeX/analysis_sections/2-motivation.tex
\section{Motivation: Eliminating Objective Mismatch in Pre-Trained Language Models}
\label{sec:motivation}

\future{Shreyas: @Pranjal Please sprinkle in appropriate citations in this section.}

Large pre-trained language models (PLMs) are massive neural networks that are trained on a huge corpus of texts using a self-supervised learning objective.
Originally utilized for representation learning~\cite{Devlin2019BERTPO, liu2019roberta} with encoder-only models, recent research, particularly influenced by \citet{Brown2020LanguageMA}, has shifted its focus towards training PLMs to directly generate answers for textual problems.
State-of-the-art PLMs typically employ an auto-regressive transformer architecture \cite{Vaswani2017AttentionIA} and are trained with a causal language modeling objective.
These models implicitly capture a conditional probability distribution $\pi_\theta$, reflecting the likelihood of sampling the next token after observing a sequence of previous tokens.
The probability of a text sequence $x := (x_1, \ldots, x_T)$, under this model is denoted as $\Pr(x; \pi_\theta) = \prod_{t = 1}^{T - 1} \pi_{\theta}(x_{t + 1} \mid x_{t}, \ldots, x_1)$.
The model is trained to estimate the pre-training data generating probability distribution over text sequences by minimizing the (forward) KL divergence between the model's data-generating distribution and the pre-training data distribution, denoted by $\Ppre(\cdot)$.
\begin{align}
    \min_{\theta} D_\textrm{KL}(\Ppre(x) \mid \mid \Pr(x; \pi_\theta)) = \min_{\theta} \mathbb{E}_{x \sim \Ppre}[\log \Ppre(x)] - \mathbb{E}_{x \sim \Ppre} [ \log \Pr(x; \pi_\theta)].
\end{align}
The first term, representing the entropy of $\Ppre$, is independent of $\theta$ and can be disregarded during optimization.
Consequently, the objective simplifies to the following cross-entropy minimization form:
\begin{align}
    \min_{\theta} - \mathbb{E}_{x \sim \Ppre}[\log \Pr(x; \pi_\theta)].
\end{align}
The expectation is approximated using samples from an unsupervised pretraining text corpus $\mathcal{D}$, which comprises text sequences sampled from $\Ppre$. This leads us to the following objective:
\begin{align}
    \min_{\theta} -\frac{1}{|\mathcal{D}|} \sum_{x \in \mathcal{D}} \sum_{t = 1}^{T - 1} \log \pi_{\theta}(x_{t+1} \mid x_t, \dots, x_1).
\end{align}

The remarkable property about PLMs lies in the contrast between the simplicity of the training recipe and the remarkable results that they deliver \cite{Brown2020LanguageMA}.
Simply capturing language statistics along with scaling up the number of trainable parameters, endows PLMs with robust semantic representations, vast commonsense knowledge, and strong pattern-following capabilities.
However, for adopting PLMs to assist humans with tasks that require an understanding of human intentions and the ability to follow instructions, the simple training recipe of PLMs is insufficient.
These models demonstrate a shallow understanding of human intentions, often generating undesirable outputs, including incorrect facts or conveying biased and toxic opinions.

Fundamentally, PLMs suffer from an \textit{objective mismatch} problem: the training-time objective of capturing language statistics does \textit{not} necessarily align with the deployment-time objective of fulfilling a human user's specific goals.
Eliminating this mismatch at first glance seems feasible: just train PLMs to optimize for the user objective.
Unfortunately, for many tasks, it is impossible to express the user objective as an optimization target.
For example, when a user's objective pertains to eliciting humorous responses, establishing specific criteria for objectively evaluating the humor in a generated response becomes an inherently challenging task.

There are currently two primary ways to deal with the problem: the behaviorist approach and the cognition-driven approach.
The behaviorist approach, implemented by supervised fine-tuning (SFT), aims to replicate observable behaviors that humans perceive as desirable without explicit consideration of the underlying user objective. For instance, if a user desires good summaries of articles, this approach trains a model to imitate examples of good summaries without explicitly defining the criteria for a good summary.
In contrast, the cognition-driven approach, implemented by reinforcement learning from human feedback (RLHF), aims to uncover the underlying user objective that governs the observed behaviors. It then updates the model by optimizing the uncovered objective.
This approach relies on certain assumptions---which in the case of RLHF are: (i) the user objective can bear the form of a reward function, which can assign a numerical score to behaviors of the model, and (ii) this function can be approximated by a machine learning model (e.g., a neural network).
RLHF estimates this reward function and updates the PLM via reinforcement learning to optimize for rewards.
Regardless of the approach, the process of addressing the objective mismatch problem is commonly referred to as the \textit{fine-tuning} or \textit{alignment} process.
Presently, state-of-the-art language models typically initiate this process with the behaviorist approach, followed by the cognition-driven approach.

\subsection{Bayesian Interpretation of RLHF}

RLHF relies on observing \textit{human feedback} to deduce the (latent) user reward function.
Human feedback is provided on the outputs from a language model.
RLHF assumes that there exists an underlying human reward function that governs the feedback they provide in a particular manner,
i.e., there exists some mapping from reward to actions of a human.
Suppose the reward function is being inferred by a model $R_{\phi}$ parameterized by $\phi$.
Adopting a Bayesian inference perspective \cite{Korbak2022RLWK}, the parameters $\phi$ can be viewed as a hypothesis with the dataset of human feedback $\mathcal{D}_\text{HF}$ as the evidence for this hypothesis.
Given a prior distribution over the hypothesis $\Pr(\phi)$, we can apply Bayes' rule to derive the posterior distribution over the hypotheses after observing the evidence as:
\begin{align}
    \Pr(\phi \mid \mathcal{D}_\text{HF}) \propto \Pr(\mathcal{D}_\text{HF} \mid R_\phi) \Pr(\phi)
\end{align}

Reward modeling in RLHF can be seen as computing the maximum a posteriori (MAP) estimate of the parameters of a reward model,
\begin{align}
    \phi_{\textrm{MAP}} = \arg\max_{\phi} ~\Pr(\phi \mid \mathcal{D}_\text{HF}) & = \arg\max_{\phi} ~\underbrace{\Pr(\mathcal{D}_\text{HF} \mid R_\phi)}_{\text{(a)}} ~\underbrace{\Pr(\phi)}_{\text{(b)}}
\end{align}

The first term (a) is the log-likelihood of the feedback dataset, specifying how a human's internal objective (reward function) governs their feedback.
The second term (b) represents constraints on the hypothesis space, which is enforced through explicit and implicit regularization techniques in neural-network training.

The presented framework raises two major questions:
\begin{enumerate}
    \item What is the form of the likelihood function $\Pr(\mathcal{D}_\text{HF} \mid R_\phi)$? In other words, how do we mathematically model the influence of a human's latent objective on their observable feedback?
    \item What is the reinforcement learning algorithm used for optimizing the reward model? In other words, how do we ensure the model acts consistently with its objective?
\end{enumerate}
A set of answers to these questions forms the basis for an RLHF algorithm.
The RLHF methodology, popularized by \citet{DeepRLHF-Christiano2017DeepRL}, employs pairwise ranking feedback and uses the Bradley-Terry model~\cite{Bradley1952RankAO} as the likelihood function.
Proximal Policy Optimization (PPO)~\cite{PPO-Schulman2017ProximalPO} is elected as the reinforcement learning algorithm.

Before we move into the analysis of this method, we urge the readers to take a moment to reflect on the choices and assumptions we have made so far to derive the general recipe of RLHF.
Are there alternative choices?
Can the assumptions be relaxed or improved?
Thinking critically about these foundational decisions is the key to understanding the strengths and weaknesses of RLHF algorithms and innovating them.
For example, the recently proposed direct preference optimization (DPO) approach~\cite{rafailov2023direct} replaces reinforcement learning with a reformulation of the objective.
Next, we formalize the problem setup of text generation as an agent interacting with a sequential decision process, laying the foundation for the analysis of RLHF.
We refer the reader to Section \ref{sec:rlhf_training} for a detailed outline of the RLHF procedure, and Figure \ref{fig:rlhf_overview} for a summarized overview.

%% file: LaTeX/analysis_sections/3-rl-formulation.tex
\section{Formulation: Text Generation as Sequential Decision-Making}
\label{sec:llm_rl_formulation}

\begin{figure}[t]
    \centering
    \includegraphics[width=0.9\columnwidth]{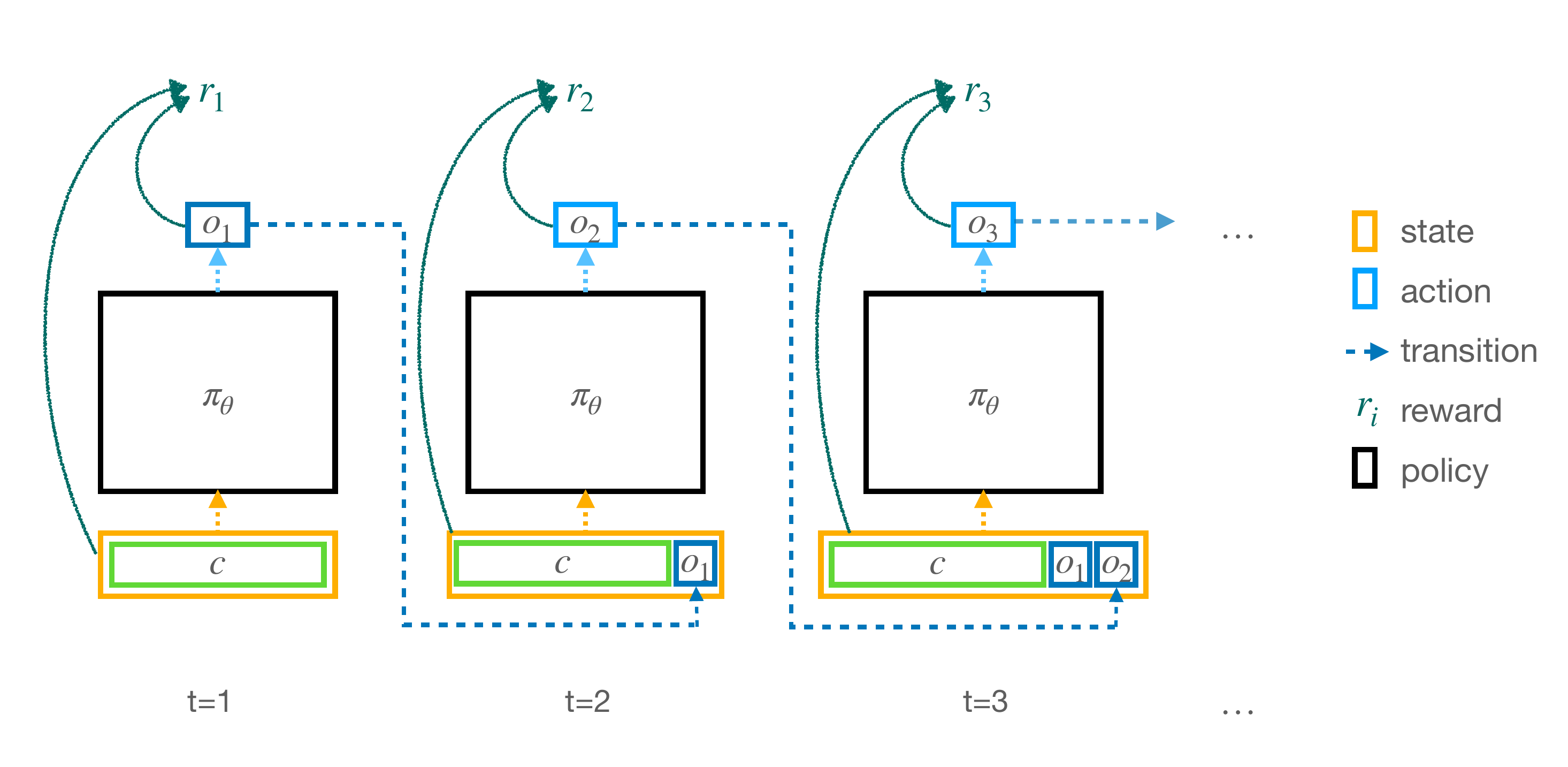}
    \caption{
        Text generation from LLMs modeled as a Markov decision process.
        The generation process is auto-regressive, utilizing the token output (action) from the previous time step and the context (state) as input to produce the next token through the language model (policy).
        Given a context $c$, the language model produces the token $o_1$ at the first timestep. A concatenation of the two $[c, o_1]$ forms the input to the policy at the next timestep (Table \ref{tbl:formulation}).
        A reward function scores the generated output for a given context.
    }
    \label{fig:output_mdp}
\end{figure}

In this section, we formulate the text generation procedure from a language model as a sequential decision-making process.
This formulation is essential for constructing reinforcement learning algorithms.

\paragraph{Markov decision process.}
A common framework for modeling sequential decision-making processes is \textit{Markov Decision Process} (MDP) \cite{markov1954theory}.
An MDP is defined as a tuple $(\mathcal{S}, \mathcal{A}, p, R, \rho)$ where $\mathcal{S}$ is the set of states, $\mathcal{A}$ is the set of actions, $p: \mathcal{S} \times \mathcal{A} \to \Delta(\mathcal{S})$ is the transition function, $R: \mathcal{S} \times \mathcal{A} \to \mathbb{R}$ is the reward function, and $\rho: \mathcal{S} \to \Delta(\mathcal{S})$ is the initial state distribution.
Each sequential time step of the process is denoted by $t$, and $s_t, a_t, r_t$ denote the values of the state, action, and reward at time step $t$.
A discounting factor $\gamma \in (0,1]$ is defined for discounting rewards over time, particularly useful for modeling an MDP with an infinite number of time steps (i.e., an infinite-horizon MDP).
However, the outputs of language models are truncated after a finite number of steps.
We use $T$ to denote the maximum time step.

An agent acts in an MDP using a policy $\pi: \mathcal{S} \rightarrow \Delta(\mathcal{A})$.
The agent starts in state $s_1 \sim \rho(\cdot)$.
At time step $t$, it chooses an action $a_t \sim \pi( ~\cdot \mid s_t)$, executes the action, transitions to a new state $s_{t + 1} \sim p( ~\cdot \mid s_t, a_t)$, and receives a reward $r_t = R(s_t, a_t)$.
The term ``Markov'' in MDP refers to the Markov property, in that the distribution over the next state $s_{t + 1}$ depends on only the current state $s_t$ and action $a_t$.

\paragraph{Language models as agents in MDP.} For simplicity, we consider text generation tasks that include only one turn of interaction between the user and the model.
We make a distinction between the text that a user inputs into the model, denoted by $c$ and referred to as the \textit{context} or the \textit{prompt}, and the text that the model generates by itself to the context, denoted by $o$ and referred to as the \textit{output} or simply the \textit{generated text}.

Let $V$ be the set of all tokens that the model can generate (the vocabulary), $\mathcal{C}$ the set of all possible contexts, and $\mathcal{O}$ the set of all possible outputs.
Given a context $c \in \mathcal{C}$ as input, the model generates an output $o \in \mathcal{O}$ token by token.
Specifically, let $o_t$ be the $t$-th token in generated output $o$, then the model parameterized by $\theta$ first outputs token $o_1 \sim \pi_{\theta}(~\cdot \mid c)$, and then conditioned on the concatenation of $o_1$ and $c$ it generates $o_2 \sim \pi_{\theta}(~\cdot \mid [c, o_1])$, and so on.

We can see that this generation process resembles an agent traversing in an MDP (Figure \ref{fig:output_mdp}).
The model acts according to a policy $\pi_{\theta}$.
The start-state distribution $\rho$ is the distribution over user-provided contexts.
The action space is the vocabulary $V$.
The action $a_t$ is the generated token $o_t$.
The state $s_t$ is the concatenation of the context $c$ and all the tokens the model has generated up to time step $t - 1$.
The transition function $p(~\cdot \mid s_t, a_t) = \delta([s_t, a_t])$ is a delta distribution, i.e., the next state is deterministic given the current state and action.
Reward $r_t$ given at time step $t$ is computed by the reward model as $R_{\phi}(s_t, a_t)$ which is either a human or a function learned from human feedback.

\begin{table}[!t]
    \centering
    \resizebox{0.8\linewidth}{!}{%
        \centering
        \begin{tabular}{lll}
            \toprule
            MDP element                  & Description                   & Text generation equivalence                              \\
            \midrule
            $s_1$                        & Initial state                 & $c$                                                      \\
            $s_t$                        & State at time step $t$        & $[c, o_{1:t-1}] = [s_1, a_{1:t-1}] = [s_{t-1}, a_{t-1}]$ \\
            $a_t$                        & Action taken at time step $t$ & $o_t$                                                    \\
            $\pi(a_t \mid s_t)$          & Policy                        & $\pi_{\theta}(o_t \mid [c, o_{1:t-1}])$                  \\
            $r_t$                        & Reward at time step $t$       & $R_{\phi}([c, o_{1:t-1}])$                               \\ %
            $\rho(s_1)$                  & Initial state distribution    & $\Pr(c)$                                                 \\
            $p(s_{t + 1} \mid s_t, a_t)$ & Transition function           & $\delta([s_t, a_t])$                                     \\
            \bottomrule
        \end{tabular}
    }
    \vspace{5pt}
    \caption{Mapping from text generation to MDP}
    \label{tbl:formulation}
\end{table}

The text generation MDP has several special properties:
\begin{enumerate}
    \item The action space is extremely large. For example, the LLaMa model~\cite{Touvron2023LLaMAOA, Touvron2023Llama2O} employs a vocabulary of size 32K. Having a gigantic action space blows up the search space for reinforcement learning algorithms.
    \item The structure of the state space is complex, as a state is essentially a text sequence. Pre-training on large amounts of texts is necessary to learn an initially good representation of this space.
    \item The initial state distribution has an enormous support. All conceivable contexts lie in the support, thus strongly testing the ability of the policy to generalize to out-of-distribution states.
    \item The reward function used for training \textit{can} differ from the evaluation reward function.
          This is because the humans providing rewards during evaluation may be different from the humans involved in trainnig the reward model. Analogous to transfer learning in RL, the agent must then adapt to the new reward function.
    \item The transition function is deterministic. Algorithmic and analysis tools tailored for deterministic MDPs can be applied.
\end{enumerate}
Thus, solving a text generation MDP requires specialized treatment that takes advantage of its properties and overcomes its inherent challenges.
Reinforcement learning \cite{sutton2018reinforcement,bertsekas1996neuro} provides solutions for optimally solving an MDP, i.e., learning a policy that maximizes the accumulated reward. Consequently, RLHF updates the language model to generate more \textit{rewarding} outputs. Naturally, the reward function plays a critical role in the process of fine-tuning model outputs, determining practical and fundamental limits \cite{casper2023open} of the efficacy of RLHF.

%% file: LaTeX/analysis_sections/4-reward-learning.tex
\section{The Role of Reward}
\label{sec:role_of_reward}

The goal of reward learning in RLHF is to convert human feedback into an optimizable reward function.
The reward serves a \textbf{dual purpose}:
it encodes the \textit{task information} (for example, identical input-output pairs would receive distinct rewards depending on whether the task involved summarization or
text expansion)
\footnote{Unless the task is specified in the input prompt itself, in which case the inputs differ.}
as well as \textit{preferences} over those outputs (a condescending summary is rewarded less than a neutral summary).
The reward thus encodes relevant information for measuring (Section \ref{sec:measuring_alignment}) as well as inducing alignment with human objectives.
By setting the reward function of the sequential decision process to the one estimated from human feedback $R_\phi$, reinforcement learning algorithms can be used to learn a language model policy that maximizes the cumulative reward, resulting in an \textit{aligned} language model.

\subsection{Oracular Reward and the Role of Human Feedback}
\label{sec:oracular_reward}

An implicit assumption made in RLHF is that a human's feedback behavior is governed by and can be represented as an \emph{oracular} reward function $R^{\star} : \mathcal{C} \times \mathcal{O} \rightarrow \mathbb{R}$.
We assume that this function is deterministic in line with the current methodology.
The function takes as input a context $c$ and an output $o$, and outputs a scalar number reflecting the preference on $o$ as a continuation of $c$.
Because the $[c, o]$ is essentially a state in the MDP formulation, the reward function is essentially defined over states of the MDP.
\textit{The language model that maximizes the oracular reward accurately reflects the goals and preferences inherent in the human feedback}, and maximization of this reward consequently aligns the model with the human preferences.
The oracular reward may not be accessible or learnable, but under the reward hypothesis \cite{sutton2004rewardhypo,silver2021reward}, the mere existence of such a reward may be assumed---though this may be challenged \cite{knox2022models}. The oracular reward forms the golden standard for training as well as evaluating any language model.

In general, humans can give a variety of feedback.
RLHF operates with feedback that discloses information about the oracular reward function.
Most methods focus on two types of feedback: \textit{point-wise numerical feedback} (or rating), and \textit{pairwise ranking feedback} (or preferences).
Providing ratings is the most straightforward way to communicate the reward function.
Given a pair $(c, o)$, the rating is a scalar $r = R^{\star}(c, o)$.
While ratings can be fed directly into a reinforcement learning algorithm, learning a reward model takes advantage of the generalizability of the reward model on unseen outputs and contexts.

Preference feedback compares two outputs generated for the same context.
Given two outputs $o$ and $o'$ generated for context $c$, a human denoted a preference $o \succ o'$ if the first input is preferred and $o' \succ o$ otherwise.
Preferences in their raw form are not compatible learning signals for reinforcement learning algorithms.
Hence, a reward model must be learned for this type of feedback.
To do so, an assumption must be made about the relationship between preferences and $R^{\star}$.
We will discuss this in more detail in the next section.
A discussion about the various methodologies used for encoding preferences can be found in Section \ref{sec:reward_model}.
An alternative approach for ranking outputs on the basis of preferences is provided by the \textit{learning-to-rank} paradigm \cite{liu2009learning}.

Using preference feedback offers several advantages compared to using ratings.
Firstly, we get more training data for the reward model.
In practice, people collect a ranking of $N$ outputs and create preference pairs~\cite{InstructGPT-Ouyang2022TrainingLM}.
Collecting $N$ ratings for $N$ outputs provides we get $N$ training points. Ranking $N$ outputs provided $N(N-1)/2$ pairwise comparisons.
Second, preferences require assigning a only relative order rather than an absolute precise score to an output; the latter task could take significantly more cognitive effort and is more prone to inconsistency.
Finally, a preference is presumably easier to provide because it offers a ``baseline'' for comparison (the worse output).
In contrast, when giving a rating, a human can rely on only the evaluation guidelines.

\textbf{A note on stochastic rewards:}
The reward function is considered to be a deterministic mapping from text to a scalar value. This amounts to averaging the preferences of all humans that provided human feedback. Moreover, it assumes that a human must always rate an input-output pair with the same score, discounting the inherent variability of human preferences. There are numerous scenarios---like personalization, in-context adaptation to ongoing dialogue, and diverse output generation---where a deterministic mapping is limiting. The rewards are more appropriately modeled as being stochastic, wherein each input-output pair is scored by a distribution over scalar rewards, say $r \sim R_\text{human}(\cdot \mid c, o)$.
This modeling accounts for the two sources of uncertainty: (i) uncertainty over the specific human from a group of humans who provide feedback, and (ii) variability in a human's preferences due to changes in unobserved factors \cite{Nguyen2017ReinforcementLF}.
Some work in reinforcement learning aims to address this by learning Bayesian preferences, primarily for uncertainty quantification and safety analysis \cite{ramachandran2007bayesian,brown2019deep}, and can be adapted to model a distribution of preferences over text.
Some recent efforts along these lines \cite{barnett2023active} have proven to be effective.
We focus on deterministic rewards for the analysis that follows.

\subsection{Reward Modeling}
\label{sec:btl_intro}

Learning a reward model serves two purposes: (i) to convert RLHF into a canonical reinforcement learning problem, and (ii) to reduce the cost of online feedback-collection.
Reinforcement learning algorithms define their objective in terms of a reward function.
To apply these algorithms, we need to infer a reward function from a feedback dataset, collecting which is notoriously expensive.
Currently, large language models require thousands to millions of feedback data points.
To gather that amount, many human evaluators need to be recruited to work in parallel.
To ensure the assumptions regarding the oracular reward function hold, the evaluators must be trained to agree with one another on the evaluation criteria.
This process is continual: multiple rounds of feedback collections need to be conducted to iteratively improve the model.
The premise of approaches that learn a reward model is that the generalization error of the reward model is expected to decrease faster than that of the policy as a function of the number of labeled data points, arising from the notion that supervised learning is often considered a simpler problem than generative modeling.

Following the previous section, we denote the reward model by $R_{\phi}(c, o)$ and the feedback dataset by $\mathcal{D}_\text{HF}$.
Our goal is to decide a likelihood function $\Pr(\mathcal{D}_\text{HF} \mid \phi)$
and find $\phi$ that maximizes this function:
\begin{align}
    \max_\phi ~\Pr(\mathcal{D}_\text{HF} \mid \phi)
\end{align}

With rating feedback, the reward-modeling problem can be formulated as a prediction problem with continuous output.
A common objective for this type of problem is the minimization of the mean squared error (MSE):
\begin{align}
    \min_{\phi} \sum_{(c, o, r) \in \mathcal{D}_\text{HF}} (R_\phi(c, o) - r)^2
\end{align}

To incorporate preference feedback, we need to choose the form of the likelihood function denoting each preference, i.e., $\Pr((o \succ o', c) \mid \phi)$.
The RLHF method of \citet{InstructGPT-Ouyang2022TrainingLM} employs the Bradley-Terry model to represent the likelihood of a data point:
\begin{align}
    \Pr((o \succ o', c) \mid \phi) & = \frac{\exp(R_\phi(c, o))}{\exp(R_\phi(c, o)) + \exp(R_\phi(c, o'))} \\
                                   & = \frac{1}{1 + \exp(R_\phi(c, o') - R_\phi(c, o))}                    \\
                                   & = \sigma[R_\phi(c, o) - R_\phi(c, o')]
\end{align} where $\sigma(x) = \frac{1}{1 + e^{-x}}$ is the sigmoid function.
The learning objective for maximizing the log-likelihood of the dataset $\mathcal{D}_\text{HF}$ is,
\begin{align}
    \max_{\phi} \sum_{(c, o, o') \in \mathcal{D}_\text{HF}} \log \Pr((o \succ o', c) \mid \phi) = \max_{\phi} \sum_{(c, o, o') \in \mathcal{D}_\text{HF}} \log \sigma[R_\phi(c, o) - R_\phi(c, o')] ~.
\end{align}
In Section \ref{sec:rlhf_reward_function}, we further generalize the form of feedback and the likelihood function to conduct an analysis independent of the specifics of particular design choices.

\subsection{Measuring Alignment}
\label{sec:measuring_alignment}

Evaluation of natural language tasks is a difficult problem, and the study of evaluation metrics is an active area of research.
Of particular importance, and difficulty, is to measure the \textit{alignment} of a language model to a human's objectives, which in practice is evaluated along the axes of helpfulness, harmlessness, and honesty.
The oracular reward that governs a human's preferences serves as a yardstick for measuring the degree of alignment.
The task of alignment is then reformulated as encoding the preferences demonstrated by a human into {a reward function}, and updating the parameters of the language model to produce output that maximizes this reward.

A reward provides an analytical metric to measure the overall performance of a language model $\pi$, where the performance captures the degree of alignment with human preferences along with the degree of satisfaction of the task itself \cite{ngo2022alignment}.
The performance of a model $\pi$, for distribution over contexts $d_C(\cdot)$, can be measured by averaging the rewards for the outputs generated by $\pi$ given the contexts. Let the performance be denoted by $J(\pi)$:
\begin{equation}
    J(\pi) := \sum_c \sum_o d_C(c) \pi(o \mid c) R^\star(c, o) = \E_{c \sim d_C(\cdot)} \left[ \E_{O \sim \pi(\cdot|c)}  \left[ R^\star(c, O) | C=c\right] \right]
    \label{eq:rlhf_eval}
\end{equation}
The context distribution $d_C(\cdot)$ can be the distribution of contexts in the training data, test data, or a held-out validation dataset, depending on the data on which the performance of the model is being evaluated.
The sequential nature of the output generation equivalently allows us to express $J(\pi)$ as:
\begin{equation}
    J(\pi) := \E_{\pi}\left[ \sum_t R^\star(s_t, a_t) \right]  = \E_{\pi}\left[ \sum_t R^\star(c, o_{1:t-1}, o_t) \right] =  \sum_t \E_{\pi}\left[ R^\star(c, o_{1:t-1}, o_t) \right]
    \label{eq:rlhf_seq_eval}
\end{equation}
In practice, most current reward models only provide a reward after the complete output has been generated and Equation \ref{eq:rlhf_seq_eval} reduces to Equation \ref{eq:rlhf_eval}. The definition of $J(\pi)$ uses the \textit{oracular} reward that is not accessible in practice. An estimate of the performance can obtained from the estimated reward $R_\phi$, by plugging it into Equation \ref{eq:rlhf_eval}:
\begin{equation}
    \widehat{J}(\pi) :=  \E_{c \sim d_C(\cdot)} \left[ \E_{O \sim \pi(\cdot|c)}  \left[ R_\phi(c, O) | C=c\right] \right]
    \label{eq:rlhf_eval_est}
\end{equation}

The pre-trained language model is denoted by $\pipre: \mathcal{C} \to \Delta(\mathcal{O})$ and the model updated using RLHF by $\pirlhf: \mathcal{C} \to \Delta(\mathcal{O})$.
The goal of RLHF is to update the parameters of $\pirlhf$ such that $J(\pirlhf) \geq J(\pipre)$, i.e., as evaluated using the oracular reward. In practice, it is only possible to verify that $\widehat{J}(\pirlhf) \geq \widehat{J}(\pipre)$, which may be non-informative when the estimated reward model $R_\phi$ has inaccuracies for the context-output pairs being evaluated (Section \ref{sec:misalignment}).

%% file: LaTeX/analysis_sections/5-estimated-reward.tex
\section{Inferring the Reward from Human Feedback}
\label{sec:rlhf_reward_function}

In the following sections, we study the properties of the reward estimated from human feedback.
As reviewed in Section \ref{sec:reward_model}, various procedures exist for encoding human feedback into a reward model.
Both the form of human feedback and the encoding mechanism continue to be studied further, with the procedures continually evolving and improving.
Currently, the most common form of human feedback is pair-wise preference feedback that is encoded into a reward according to the Bradley-Terry model (Section \ref{sec:btl_intro}).
To perform an analysis agnostic to specifics of a particular reward learning method,
\begin{enumerate}
    \item Let \texttt{feedback} denote a general form of sufficiently informative feedback.
    \item Let $\Omega$ denote the \textit{model of human behavior}, or the encoding mechanism, that maps the feedback and the text to a reward value.
\end{enumerate}
The generality of this formulation allows the following analysis to cover all existing RLHF-style approaches (for example, RLAIF \cite{Bai2022ConstitutionalAH}) as well as future methods for fine-tuning LLMs, that employ a reward model.

Let $\mathcal{D} := \{ (c, o): c \in \mathcal{C}, o \in \mathcal{O} \}$ denote a hypothetical dataset of all possible contexts and outputs that a language model can encounter, i.e., a humongous dataset of size $|\mathcal{C}| \times |\mathcal{O}| = |V|^{T}$. This dataset cannot be realized in practice and is invoked to shed light on the practical limitations of the existing methodology.
Denote the dataset of collected human feedback by $\mathcal{D}_{\text{HF}} := \{ (c, o, \text{\texttt{feedback}}): c \in \mathcal{C_{\text{HF}}}, o \in \mathcal{O_{\text{HF}}} \}$ where $\mathcal{C_{\text{HF}}} \subset \mathcal{C}, \mathcal{O_{\text{HF}}} \subset \mathcal{O}$ are the subsets of context-output pairs (human-)annotated with \texttt{feedback}.\footnote{The subsets are significantly smaller than $\mathcal{C}$ and $\mathcal{O}$. Additionally, the feedback can be of any form: ratings, pair-wise feedback, or language feedback (Section \ref{sec:human_feedback}).}
The reward encoding mechanism that maps context-output pairs along human feedback to rewards (for instance, the Bradley-Terry model) is denoted by $\Omega: (c, o, \text{\texttt{feedback}}) \to \mathbb{R}$.\footnote{%
    $\texttt{feedback}$ is overloaded to capture additional mechanism-specific meta-data.
    For instance, for pair-wise preference, $\texttt{feedback}$ can store the preference relation and the $(c,o)$ pair compared against.}

To uncover $R^\star$, it is \textit{assumed} that $\Omega$ accurately maps back human feedback to the oracular reward, i.e., for \textit{sufficiently informative} \texttt{feedback}, we have $$\Omega(c,o,\texttt{feedback}) = R^\star(c,o), ~\forall ~c,o \in \mathcal{C_\text{HF}},\mathcal{O_\text{HF}}.$$
Under that assumption, $\Omega$ can operate on $\mathcal{D}_{\text{HF}}$ to create a dataset of context-output-reward tuples,
$\mathcal{D}_{\text{rew}} = \{ (c,o,r): c \in \mathcal{C_{\text{HF}}}, o \in \mathcal{O_{\text{HF}}} \}$ where $r = R^\star(c,o)$.
With $\mathcal{D}_{\text{rew}}$, learning the reward model $R_\phi$ reduces to a regression problem employing a function approximator.
The regression problem is however \textit{underdetermined} \cite{bishop2006pattern}, and consequently multiple $R_\phi$ functions can perfectly fit the training data $\mathcal{D}_{\text{rew}}$. However, almost all of these functions fail to accurately represent the oracular reward (Figure \ref{fig:reward_generalization}).
Due to the cost of human annotation, practically human feedback can be collected on a very small subset of context and output pairs, i.e., $\mathcal{C_{\text{HF}}}, \mathcal{O_{\text{HF}}} \subset \mathcal{C}, \mathcal{O}$.
The size of the reward and feedback datasets relative to the hypothetical dataset of all possible inputs and outputs $\mathcal{D}$ can be measured by:
\begin{enumerate}
    \item Context coverage: $\kappa := \frac{| \mathcal{C_{\text{HF}}} |}{| \mathcal{C} |} $
    \item Output coverage: $\rho := \frac{|\mathcal{O}_\text{HF}|}{|\mathcal{O}'|}$, where $\mathcal{O}' = \{o: (c,o) \in \mathcal{D}, ~\forall c \in \mathcal{C}_\text{HF}\}$

\end{enumerate}
Well-understood results in supervised learning suggest that
the ratios $\rho$ and $\kappa$ along with the generalization capabilities of the function approximator \cite{nakkiran2019deep,schaeffer2023double,bailly2022effects} determine the generalization performance of the reward model for $(c,o) \in \mathcal{C}, \mathcal{O}$.
In practice, the values of $\rho$ and $\kappa$ are extremely small and consquently the reward model often incorrectly generalizes on unseen (out-of-distribution) context-output pairs, assigning incorrect rewards to such inputs.
In the following sections, we study practical limitations of estimating reward models.

\subsection{Limitations of the Reward Model}
\label{sec:analysis_reward_model}

The reward model parameterized ${R}_\phi: \mathcal{C} \times \mathcal{O} \to \mathbb{R}$
is trained on $D_{\text{rew}}$ using a sufficiently representative function approximator, to perfectly fit the training data, that is $R_\phi(c, o) = R^\star(c, o), ~\forall ~o,c \in D_{\text{rew}}$.
The limitations of the resultant reward model may be studied under the following categories:

\begin{figure}[t]
    \centering
    \includegraphics[width=0.8\columnwidth]{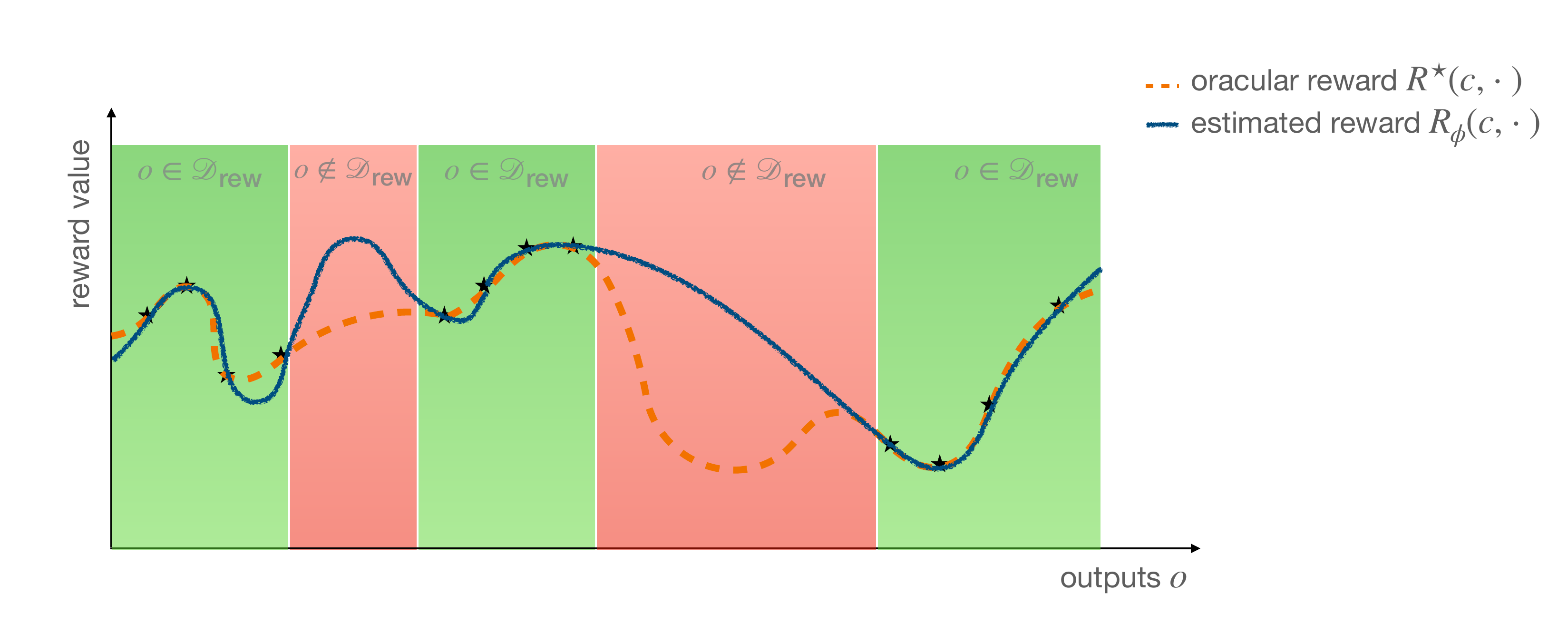}
    \caption{
    The reward model tends to misgeneralize for inputs not found in its training data, i.e., for $(c, o) \notin \mathcal{D}_{\text{rew}}$. This occurs in two ways: 1) when the context is not sampled by the prompting distribution for generating output and receiving feedback on (represented by $\kappa$), and 2) when the support of the output generating distribution---the language model---for a context does not span all possible outputs (represented by $\rho$). The latter is depicted in this figure.
    }
    \label{fig:reward_generalization}
\end{figure}

\textbf{Misgeneralization:}
Human feedback is obtained on a very small subset of all possible context-output pairs.
This partial coverage over contexts and outputs in $D_{\text{rew}}$ combined with the use of function approximators for learning the reward model results in the reward model $R_\phi(c,o)$ incorrectly generalizing to data points that are \textit{out-of-distribution} relative to $D_{\text{rew}}$.
We have assumed a sufficiently representative function approximator that perfectly fits the training data,
$\E_{c, o \sim \mathcal{D}_\text{rew}}\left[(R^\star(c, o) - R_\phi(c,o))^2\right] = 0$.
However, it cannot be ensured that
$\E_{c, o \notin \mathcal{D}_\text{rew}}\left[(R^\star(c, o) - R_\phi(c,o))^2\right]$
will be zero. It would require a function approximator to perfectly generalize outside the training data distribution, which is not generally attainable, especially when the ratios $\rho, \kappa$ are minuscule.

\textit{The benefits of reinforcement learning algorithms over other methods
    for finetuning
    are contingent on access to an accurate reward function (Section \ref{sec:why_rl}), necessitating accurate
    out-of-distribution generalization of the reward model.}

The inaccurate extrapolation out-of-distribution
results in an `imperfect' reward model that provides feedback on context-output pairs in a manner that when optimized, arbitrarily misaligns with human feedback (and resultant preferences) for those context-output pairs.
The output distribution of $\pirlhf$ trained on this inaccurate feedback and can only be as good (or bad) as the reward signal provided by the reward model.
This \textit{inaccurate generalization in the reward model} is one of the primary causes of phenomena like `reward hacking' and `hallucinations' \cite{kalai2023calibrated}, observed in practice.

\textbf{Delayed feedback and Reward Sparsity:}
Reinforcement learning algorithms benefit from dense rewards as they serve to quickly guide the agent to rewarding states, providing informative feedback to intermediate actions along the trajectory.
In RLHF, the feedback from human annotators is obtained for \textit{complete} output generations. Consequently, the reward model is trained to provide reward feedback only at the end of the generated output for a given context.
This delayed feedback increases the difficulty of optimization with RL algorithms, increasing their sample complexity.
Sparse feedback is a constraint inherent to dealing with text and language \cite{Sokolov2016BanditSP}, as it is often unlikely for a human to provide feedback on incomplete sentences.
Methods in RL developed to deal with sparse feedback, for instance by stitching together information from partial trajectories \cite{andrychowicz2017hindsight}, cannot be applied directly to textual output due to the semantic constraints of dealing with partial sentences.
Denser rewards and corresponding feedback result in faster training, improved sample efficiency \cite{wu2023fine}, and potentially better generalization.
Insights from linguistics may be employed to obtain feedback on partial output generations and in turn denser rewards.

\textbf{Marginalization over preferences:}
The reward model averages over the preferences of all human annotators (and other sources of feedback) to output a deterministic scalar reward for a given context-output pair.
The expectation is that averaging over the preferences of multiple sources would be representative of the preferences of an {average human persona} \cite{deshpande2023anthropomorphization}.
The results in rewards that are inconsistent with any single human's preferences.
Such preferences are more appropriately denoted by an \textit{distribution} of rewards for a context-output pair.
A deterministic model, in addition to discounting the uncertainty and variability of human preferences, cannot model such a distribution, highlighting a case of \textit{model misspecification}.

The reward model forms the core component of RLHF and dictates the performance of a language model.
The aforementioned shortcomings of the reward model highlight the need for safety measures that must be employed while using a language model fine-tuned using RLHF.

%% file: LaTeX/analysis_sections/6-rl-on-reward.tex
\section{Reinforcement Learning with Imperfect Rewards}

Reinforcement learning algorithms can be broadly categorized into value-based methods and policy-gradient methods \cite{sutton2018reinforcement}. Value-based methods aim to learn the \textit{value} of states (or state-action pairs) as measured by the expected cumulative future reward from that state under a policy.
These values serve as guides for picking the most rewarding actions from each state, and the policy can be inferred from the value functions. Policy-gradient methods directly train a parameterized policy using reward feedback to perform gradient ascent over the policy parameters and maximize the expected cumulative reward. A benefit of policy gradient methods for language tasks is that they naturally permit the optimization of stochastic policies, making them amendable for optimizing language models with stochastic decoding algorithms. Below we provide a brief overview of policy gradient algorithms and refer interested readers to \citet{sutton1999policy,williams1992simple,weng2018PG} for a more rigorous treatment of the topic.

\subsection{Policy Gradient Algorithms}

Policy gradient algorithms update the parameters of an agent's policy using reward feedback.
Being gradient-based algorithms, their update rule is of the form:
\begin{equation}
    \theta \longleftarrow \theta + \alpha \nabla_\theta J(\pi_\theta)
    \label{eq:gen_pg}
\end{equation}
where $J(\pi_\theta)$ is the performance (Equation \eqref{eq:rlhf_eval}) of the policy parameterized by $\theta$.
The gradient of the performance of a policy $\nabla_\theta J(\pi_\theta)$ can be estimated from samples in numerous ways, each affording varying degrees of variance and estimation error.
In sparse rewards settings, the gradient estimation variance is a common issue that baselines \cite{mei2022role} help address.
A class of methods called \textit{actor-critic methods}
update the policy by leveraging estimated value functions, called critics, to reduce gradient estimation variance.
The algorithm used for training most current state-of-the-art large language models, Proximal Policy Optimization (PPO) \cite{PPO-Schulman2017ProximalPO} is an actor-critic algorithm with improvements over vanilla actor-critic to ensure stability during training.
The improvements restrict parameter updates at each iteration to prevent the policy distribution from drastically changing.
The training loss objective for PPO (PPO-Clip) takes the form:
\begin{equation}    \mathcal{L}_{\text {ppo-clip }}(\theta)
    =
    \mathbb{E} \left[\min \left(\frac{\pi_\theta\left(a_t \mid s_t\right)}{\pi_{\theta_{\text {old }}}\left(a_t \mid s_t\right)} \hat{A}(s_t, a_t), \operatorname{clip}\left(\frac{\pi_\theta\left(a_t \mid s_t\right)}{\pi_{\theta_{\text {old }}}\left(a_t \mid s_t\right)}, 1-\epsilon, 1+\epsilon\right) \hat{A}(s_t, a_t)\right)\right]
    \label{eq:ppo}
\end{equation}
where $\hat{A}(s_t, a_t)$ is the estimate of the \textit{advantage function} $A(s_t, a_t) := Q(s_t, a_t) - V(s_t)$ that captures the \textit{advantage} obtained in terms of cumulative reward by taking an action $a_t$
from state $s_t$ and then following the current policy, relative to following the policy starting from state $s_t$. While this background suffices for the discussion in this paper, we urge the reader to refer to \citet{weng2018PG} for a more in-depth explanation of the topic.

\subsection{Misalignment due to Imperfect Rewards}
\label{sec:misalignment}

In practice, a KL penalty $D_\text{KL}\left( \pi_\theta || \pipre \right)$ with some weight $\beta$ is added to the PPO training objective.
This can be interpreted either as a regularizer or a prior which helps prevent overoptimization of an imperfect reward model.
Using a reward model $R_\phi$,
the policy at convergence learnt by training with the updated PPO objective can expressed directly as a function of the reward \cite{ScaleLanguageFeedback-Scheurer2023TrainingLM,rafailov2023direct} as,
\begin{equation}
    \pirlhf (o \mid c) \propto \pipre(o \mid c) \exp\left(\frac{1}{\beta} R_\phi(c,o) \right)
    \label{eq:imperfect_policy}
\end{equation}
where $\beta$ is the weight on the KL penalty.
Let $\mathcal{C}_\text{rew} \subset \mathcal{C}$ be the set of contexts in $\mathcal{D}_\text{rew}$.\footnote{Note that $\mathcal{C}_\text{rew}$ is the same as $\mathcal{C}_\text{HF}$. The subscript is used for clarity under the current context.}
After training, $\pirlhf$ must generate desirable (most rewarding) outputs when prompted with $c \in \mathcal{C}_\text{rew}$.
But for out-of-distribution contexts, where the reward estimation may be erroneous, the output distribution of $\pirlhf$ may be arbitrarily misaligned with human preferences and generate undesirable output.
This \textit{misalignment} can be quantified by comparing against the policy trained with the oracular reward.
The set of contexts on which the performance of $\pirlhf$ is evaluated is denoted by $\mathcal{C}_\text{eval}$ with $d_{\mathcal{C}_\text{eval}}$ being the distribution over those contexts. Let $C' = \mathcal{C}_\text{eval} / \mathcal{C}_\text{rew}$ be the set of contexts in the evaluation set that are not present in $\mathcal{C}_\text{rew}$. The performance of $\pirlhf$ is given by:

\[\begin{split}
        J(\pirlhf) &= \E_{c \sim d_{\mathcal{C}_\text{eval}}(\cdot), ~o \sim \pirlhf(\cdot \mid c)} \left[ R^\star(c, o)\right] \\
        &= \sum_{\substack{{\color{teal} c \in \mathcal{C}_\text{rew}}, \\ o \in \mathcal{O}}} d_{\mathcal{C}_\text{eval}}(c) \pirlhf(o \mid c) R^\star(c, o) + \sum_{\substack{{\color{red} c \in C'}, \\~o\in \mathcal{O}}} d_{\mathcal{C}_\text{eval}}(c) \pirlhf(o \mid c) R^\star(c, o) \\
        &\overset{(a)}{=} \sum_{\substack{{\color{teal} c \in \mathcal{C}_\text{rew}}, \\ o \in \mathcal{O}}} d_{\mathcal{C}_\text{eval}}(c) \pipre(o \mid c) \left[ \frac{\pirlhf(o|c)}{\pipre(o|c)} R^\star(c, o)\right]   + \sum_{\substack{{\color{red} c \in C'}, \\~o\in \mathcal{O}}} d_{\mathcal{C}_\text{eval}}(c) \pipre(o \mid c) \left[\frac{\pirlhf(o|c)}{\pipre(o|c)} R^\star(c, o)\right] \\
        &\overset{(b)}{\propto} {\small \sum_{\substack{{\color{teal} c \in \mathcal{C}_\text{rew}}, \\ o \in \mathcal{O}}} d_{\mathcal{C}_\text{eval}}(c) \pipre(o \mid c) \left[ \exp \left(\frac{1}{\beta}{\color{blue}R^\star(c,o)}\right) R^\star(c, o)\right] +   \underbrace{\sum_{\substack{{\color{red} c \in C'}, \\~o\in \mathcal{O}}} d_{\mathcal{C}_\text{eval}}(c) \pipre(o \mid c) \left[ \exp\left(\frac{1}{\beta}{\color{orange}R_\phi(c,o)}\right) R^\star(c, o)\right]}_{\text{out-of-distribution}}}
    \end{split}
\]

where (a) is permitted
by the following: $\forall o, c: \pirlhf(o|c) > 0, \pipre(o|c) > 0$ and (b) follows from Equation \eqref{eq:imperfect_policy}.
Let $\pi^*$ be the policy trained using the oracular reward with RLHF. It can be expressed as:
\[
    \pi^*_\text{rlhf} (o \mid c) \propto \pipre(o \mid c) \exp\left(\frac{1}{\beta} R^\star(c,o) \right)
\]
The performance of $\pi^*_\text{rlhf}$ can be written as:
\[
    \begin{split}
        J(\pi^*_\text{rlhf}) &= \E_{c \sim d_{\mathcal{C}_\text{eval}}(\cdot), ~o \sim \pi^*_\text{rlhf}(\cdot \mid c)} \left[ R^\star(c, o)\right] \\
        &\propto \sum_{\substack{{\color{teal} c \in \mathcal{C}_\text{rew}}, \\ o \in \mathcal{O}}} {\small d_{\mathcal{C}_\text{eval}}(c) \pipre(o \mid c) \left[ \exp \left(\frac{1}{\beta}{\color{blue}R^\star(c,o)}\right) R^\star(c, o)\right] +  \underbrace{\sum_{\substack{{\color{red} c \in C'}, \\~o\in \mathcal{O}}} d_{\mathcal{C}_\text{eval}}(c) \pipre(o \mid c) \left[ \exp\left(\frac{1}{\beta}{\color{blue}R^\star(c,o)}\right) R^\star(c, o)\right]}_{\text{out-of-distribution}}}\\
    \end{split}
\]

The performance gap $\Delta J := |J(\pi^*_\text{rlhf}) - J(\pirlhf)|$ caused by the imperfections in the reward model can be quantified as,
\[\Delta J \propto \sum_{{\color{red} c \in C'}, ~o \in \mathcal{O}} d_{\mathcal{C}_\text{eval}}(c) \pipre(o \mid c) \left[ \lvert \biggl( \exp\left(\frac{1}{\beta}{\color{blue}R^\star(c,o)}\right) -  \exp\left(\frac{1}{\beta}{\color{orange}R_\phi(c,o)}\right) \biggr) R^\star(c, o) \big\rvert\right] \]

For out-of-distribution contexts and outputs, the reward model is known to misgeneralize. The performance gap increases with increasing discrepancy from the oracular reward, and the discrepancy is further weighted by the likelihood of that $(c,o)$ pair and its oracular reward value.
Some observations from the above analysis:
\begin{itemize}
    \item $\pirlhf$ assigns high probability to highly rewarding outputs (Equation \eqref{eq:imperfect_policy}), which is beneficial in-distribution contexts but can be harmful for out-of-distribution contexts when the reward model is erroneous.
    \item The deviation of the estimated reward from the oracular reward on unseen contexts exacerbates misalignment, which can be mitigated by increasing the weight on the KL penalty due to the $1/\beta$ dependence in the exponent.
    \item However, there is a trade-off. Increasing the value of $\beta$ results in $\pi^*_\text{rlhf}$ and $\pirlhf$ being closer to $\pipre$ and have a lowered performance---due to increased weight in the KL penalty.
\end{itemize}

\subsection{Why use Reinforcement Learning Algorithms?}
\label{sec:why_rl}

The efficacy of RLHF heavily relies on the quality of the reward model, and thus a large fraction of future research must focus on improving the reward model.
Before allocating resources to that effort, it is essential to evaluate the merits and downsides of employing reinforcement learning as the fine-tuning paradigm.
In comparison to supervised learning as an alternative approach, examining the gradient updates of a (vanilla) policy gradient algorithm alongside those of a supervised learning algorithm (such as supervised fine-tuning) offers some insights.

\paragraph{Comparing Update Rules of Supervised Fine-Tuning and RLHF:}

In supervised fine-tuning (SFT), supervision is provided with positive samples, and the language model is updated to increase the likelihood of those samples under the model. Notably, there is no supervision provided for neutral or undesirable outputs, although it is a feasible option.
Given the optimal policy $\pi^*$ (which may be a human expert), the objective of SFT is,
\[\max_\theta  \E_{c \sim d_C, {\color{red} o_w \sim \pi^*}(\cdot|c)} \big[ \ln \pi_\theta({\color{red}o_w} | c) \big]\]
and thus the gradients used to update the parameters of the language model are of the form:
\[ \nabla_\theta := \E_{c \sim d_C, {\color{red} o_w \sim \pi^*}(\cdot|c)} \big[ \nabla_\theta \ln \pi_\theta({\color{red}o_w} | c) \big]. \]
This is analogous to behavior cloning in RL \cite{pomerleau1988alvinn} which is known to struggle when faced with out-of-distribution inputs.

The primary benefit that reinforcement learning algorithms provide is that they allow the language model to \textit{explore} the output space.
Through its decoding algorithm, the language model exercises control over the distribution of outputs on which feedback is acquired.
This facilitates learning from both positive as well as negative feedback,
i.e.,
\[ \max_\theta \E_{c \sim d_C, {\color{cyan} o \sim \pi_\theta}(\cdot|c)} \big[ R^\star(c,{\color{cyan}o}) \big]\]
and the (vanilla) policy gradient update is:
\[ \nabla_\theta :=  \E_{c \sim d_C, {\color{cyan} o \sim \pi_\theta}(\cdot|c)} \big[ R^\star(c,{\color{cyan}o})  \nabla_\theta \ln \pi_\theta ({\color{cyan}o}|c)  \big]~.  \]
As highlighted in color, in SFT, the gradient is estimated only from the positive samples, while in RL, it is computed for all samples (positive, negative, or neutral) \textit{weighted}  by their corresponding rewards.
The gradient updates in RL are more informative, leading to better generalization for the language model and improved sample efficiency.
Beyond exploration and richer gradients, the field of inverse reinforcement learning provides a natural formulation for training a language model with human feedback \cite{arora2021survey}.

In the following sections, we present a review of works that lead up to and are being rapidly added to this active area of research.
This review provides context for the first half of this work and also serves as a comprehensive introduction for readers interested in getting started and understanding the topic of RLHF for language models.

%% file: LaTeX/survey.tex
\section{Review of Reinforcement Learning from Human Feedback for Language Models}

\setcitestyle{authoryear}
\bibpunct{[}{]}{;}{a}{}{,}
\renewcommand{\cite}[1]{\citep{#1}}

\input{LaTeX/survey_sections/100_pretrained_llms}

\input{LaTeX/survey_categories_small}

\input{LaTeX/survey_sections/200_rlhf_overview2}

\input{LaTeX/survey_sections/210_human_feedback}

\input{LaTeX/survey_sections/220_initial_policy}

\input{LaTeX/survey_sections/230_reward_modelling}
\input{LaTeX/survey_sections/240_finetuning}

\input{LaTeX/survey_sections/250_rlhf_algos}
\input{LaTeX/survey_sections/260_properties_of_rlhf_models}

\input{LaTeX/survey_sections/350_sparse_rewards_in_rl}
\input{LaTeX/survey_sections/400_reward_alternatives}

%% file: LaTeX/survey_sections/100_pretrained_llms.tex
\subsection{Language Model Pre-Training: Foundation for Large Language Models}
\label{sec:pretrained_llms}

Language Models (LMs) have gained significant attention in recent years due to their impressive abilities to model language and retain textual knowledge. The Transformer architecture, characterized by its use of self-attention mechanisms, has become the standard for LMs~\cite{Vaswani2017AttentionIA}. It is employed in a range of models, including BERT, T5, LLaMA, GPT-3, PALM, GLaM~\cite{Devlin2019BERTPO, Raffel2019ExploringTL, Touvron2023LLaMAOA, Brown2020LanguageMA, Chowdhery2022PaLMSL, Du2021GLaMES}.

Pre-training has played an important role in the development of Large Language Models (LLMs), significantly contributing to their remarkable performance across a myriad of downstream tasks~\cite{Brown2020LanguageMA, Chowdhery2022PaLMSL, Zhang2022OPTOP}.
This process involves training models with an unsupervised training objective on extensive datasets, often comprised of a diverse mix of web content, literary works, scientific documents, and code repositories~\cite{Rae2021ScalingLM, Xie2023DoReMiOD}. The scale of these datasets is critical, with studies highlighting the superior performance of smaller models trained on larger datasets~\cite{Kaplan2020ScalingLF, Hoffmann2022TrainingCL, Touvron2023LLaMAOA}. In addition to scale, the quality of training data, ensured through deduplication and filtering of low-quality content, is a key determinant of model performance~\cite{Rae2021ScalingLM, Du2021GLaMES, Hernandez2022ScalingLA, Lee2021DeduplicatingTD}. Masked Language Modeling (MLM)~\cite{Devlin2019BERTPO} and Causal Language Modeling~\cite{Radford2018ImprovingLU} are the most common objectives used for pretraining, with latter showing notable success in recent Large Language Model series such as GPT, PaLM, OPT~\cite{Anil2023PaLM2T, OpenAI2023GPT4TR, Zhang2022OPTOP}.

Studies demonstrate that pre-training by itself is responsible for the bulk of the observed capabilities even in downstream tasks~\cite{Brown2020LanguageMA, Raffel2019ExploringTL}.
The simple pre-training objective of next, or masked, token prediction imbibes the LMs with a range of capabilities.
They are few-task learners, without the need for fine-tuning. This applies to a variety of tasks from text generation, reasoning, question answering, summarization, and translation to name a few.
However, though scaling pretrained language models (PLMs) exhibit remarkable performance across a variety of tasks,
they suffer from several limitations, such as the inability to follow human instructions~\cite{InstructGPT-Ouyang2022TrainingLM}.
\future{Refine next line}
This is because PLMs suffer from objective mismatch problems (See Section~\ref{sec:motivation}), as they are trained on generic internet data. As a result, PLMs need to learn to mimic the conflicting behavior of billions of humans. 
Further, the Maximum Likelihood Estimate on the next token prediction for such data doesn't explicitly penalize the model for hallucinating concepts, i.e., generating concepts not encapsulated within its internal representation, and even important \& unimportant errors are given equal weightage.
Moreover, pretrained models often show unintended behavior such as generating harmful, biased, untruthful, and low-quality content~\cite{RedTeaming-Perez2022RedTL}.

\paragraph{Supervised-Finetuning}

To address the shortcomings faced by PLMs, a straightforward approach is to fine-tune them on a set of high-quality downstream datasets that are indicative of the intended task and behavior. For example, for instruction-following, human annotations can be collected on a set of input prompts, or input instances of existing public datasets can be re-formatted for instruction-following format. The model is then simply fine-tuned on these human demonstrations, often with the same pretraining objective.
This increases the likelihood of generating desirable text and makes the model less biased and harmful.
Nonetheless, in order to generate high-quality text, it is crucial to note that the task of distinguishing between high and low-quality text is inherently subjective and challenging, with end users being humans. Thus, quality assessment rests on human judgment and varies significantly based on the individual evaluator's perspective~\cite{Yi2019TowardsCA, Fan2022NanoNH, FinetuningRLHF-Ziegler2019FineTuningLM}. Incorporating human feedback into such a process can be challenging, and collecting high-quality human demonstrations can be expensive and not scalable.

%% file: LaTeX/survey_categories_small.tex
\tikzstyle{my-box}=[
rectangle,
draw=hidden-draw,
rounded corners,
text opacity=1,
minimum height=1.5em,
minimum width=5em,
inner sep=2pt,
align=center,
fill opacity=.5,
]
\tikzstyle{leaf}=[my-box, minimum height=1.5em,
fill=hidden-orange!60, text=black, align=left,font=\scriptsize,
inner xsep=2pt,
inner ysep=4pt,
]
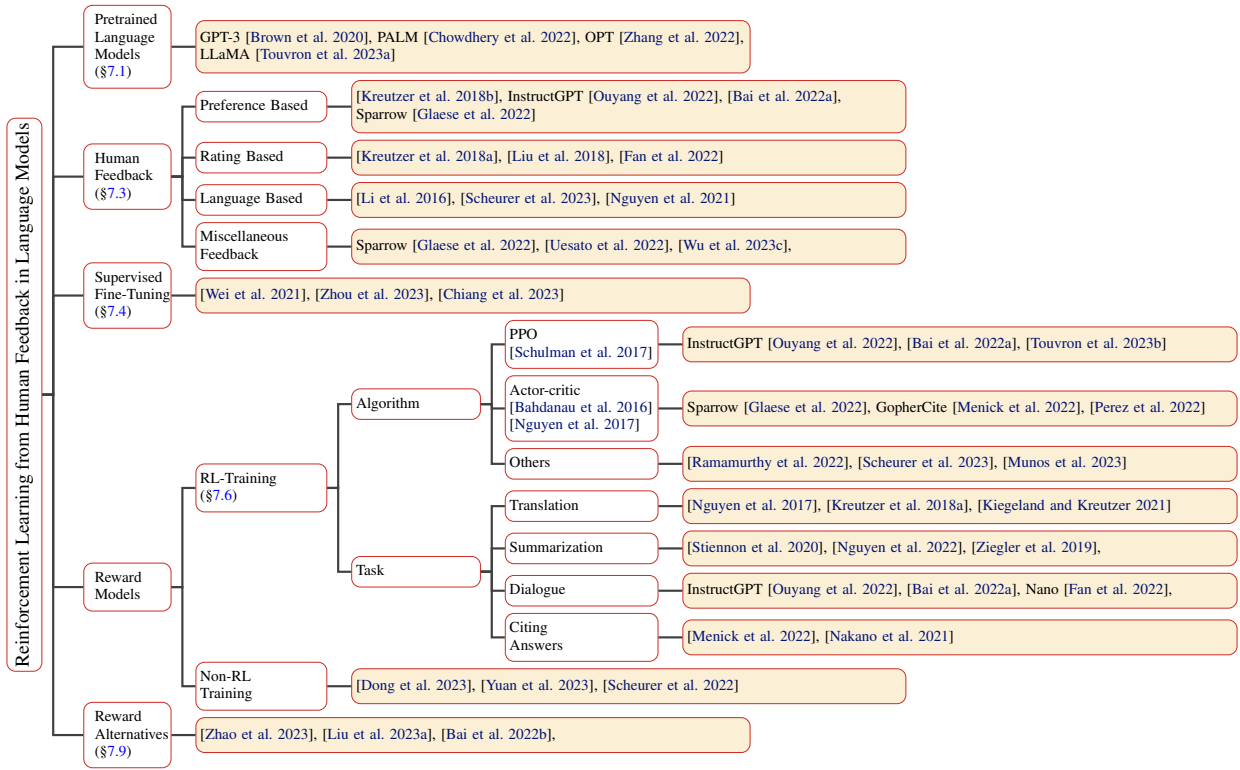
\begin{figure*}[tp]
    \centering
    \resizebox{\textwidth}{!}{
        \begin{forest}
            forked edges,
            for tree={
            grow=east,
            reversed=true,
            anchor=base west,
            parent anchor=east,
            child anchor=west,
            base=left,
            font=\small,
            rectangle,
            draw=hidden-draw,
            rounded corners,
            align=left,
            minimum width=4em,
            edge+={darkgray, line width=1pt},
            s sep=3pt,
            inner xsep=2pt,
            inner ysep=3pt,
            ver/.style={rotate=90, child anchor=north, parent anchor=south, anchor=center},
            },
            where level=1{text width=3em,font=\scriptsize,}{},
            where level=2{text width=5.6em,font=\scriptsize,}{},
            where level=3{text width=5.5em,font=\scriptsize,}{},
            where level=4{text width=6.6em,font=\scriptsize,}{},
            [
                    Reinforcement Learning from Human Feedback in Language Models, ver
                        [
                            Pretrained \\ Language \\ Models \\ (\S \ref{sec:pretrained_llms})
                            [
                                    GPT-3~\cite{Brown2020LanguageMA}{,}
                                    PALM~\cite{Chowdhery2022PaLMSL}{,}
                                    OPT~\cite{Zhang2022OPTOP}{,} \\
                                    LLaMA~\cite{Touvron2023LLaMAOA}
                                    , leaf, text width=25em
                                ]
                        ]
                        [
                            Human \\ Feedback \\ (\S \ref{sec:human_feedback})
                            [
                                    Preference Based
                                        [
                                            \cite{Kreutzer2018ReliabilityAL}{,}
                                            InstructGPT~\cite{InstructGPT-Ouyang2022TrainingLM}{,}
                                            \cite{HHHAssistant-Bai2022TrainingAH}{,} \\
                                            Sparrow~\cite{Sparrow-Glaese2022ImprovingAO}
                                            , leaf, text width=25em
                                        ]
                                ]
                                [
                                    Rating Based
                                        [
                                            \cite{Kreutzer2018CanNM}{,}
                                            \cite{Liu2018DialogueLW}{,}
                                            \cite{Fan2022NanoNH}
                                            , leaf, text width=25em
                                        ]
                                ]
                                [
                                    Language Based
                                        [
                                            \cite{DialogueHITL-Li2016DialogueLW}{,}
                                            \cite{ScaleLanguageFeedback-Scheurer2023TrainingLM}{,}
                                            \cite{nguyen2021iliad}, leaf, text width=25em
                                        ]
                                ]
                                [
                                    Miscellaneous \\ Feedback
                                        [
                                            Sparrow~\cite{Sparrow-Glaese2022ImprovingAO}{,}
                                            \cite{DeepMindSolvingMath-Uesato2022SolvingMW}{,}
                                            \cite{{FineGrainedRLHF-Wu2023FineGrainedHF}}{,}
                                            , leaf, text width=25em
                                        ]
                                ]
                        ]
                        [
                            Supervised \\ Fine-Tuning \\ (\S \ref{sec:inital_policy})
                            [
                                    \cite{Wei2021FinetunedLM}{,}
                                    \cite{zhou2023lima}{,}
                                    \cite{vicuna2023}
                                    , leaf, text width=25em
                                ]
                        ]
                        [
                            Reward \\ Models
                                [
                                    RL-Training \\ (\S \ref{sec:rlhf_training})
                                    [
                                            Algorithm \\ 
                                            [
                                                    PPO \\ \cite{PPO-Schulman2017ProximalPO}
                                                    [
                                                        InstructGPT~\cite{InstructGPT-Ouyang2022TrainingLM}{,}
                                                        \cite{HHHAssistant-Bai2022TrainingAH}{,}
                                                        \cite{Touvron2023Llama2O}
                                                        , leaf, text width=25em
                                                    ]
                                                ]
                                                [
                                                    Actor-critic \\ \cite{A2C-Bahdanau2016AnAA} \\ \cite{Nguyen2017ReinforcementLF}
                                                    [
                                                        Sparrow~\cite{Sparrow-Glaese2022ImprovingAO}{,}
                                                        GopherCite~\cite{GopherCite-Menick2022TeachingLM}{,}
                                                        \cite{RedTeaming-Perez2022RedTL}
                                                        , leaf, text width=25em
                                                    ]
                                                ]
                                                [
                                                    Others
                                                        [
                                                            \cite{Ramamurthy2022IsRL}{,}
                                                            \cite{ScaleLanguageFeedback-Scheurer2023TrainingLM}{,}
                                                            \cite{Munos2023NashLF}
                                                            , leaf, text width=25em
                                                        ]
                                                ]
                                        ]
                                        [
                                            Task
                                                [
                                                    Translation
                                                        [
                                                            \cite{Nguyen2017ReinforcementLF}{,}
                                                            \cite{Kreutzer2018CanNM}{,}
                                                            \cite{Kiegeland2021RevisitingTW}
                                                            , leaf, text width=25em
                                                        ]
                                                ]
                                                [
                                                    Summarization
                                                        [
                                                            \cite{SummarizeHF-Stiennon2020LearningTS}{,}
                                                            \cite{InteractiveRLHF-Nguyen2022MakeTM}{,}
                                                            \cite{FinetuningRLHF-Ziegler2019FineTuningLM}{,}
                                                            , leaf, text width=25em
                                                        ]
                                                ]
                                                [
                                                    Dialogue
                                                        [
                                                            InstructGPT~\cite{InstructGPT-Ouyang2022TrainingLM}{,}
                                                            \cite{HHHAssistant-Bai2022TrainingAH}{,}
                                                            Nano~\cite{Fan2022NanoNH}{,}
                                                            , leaf, text width=25em
                                                        ]
                                                ]
                                                [
                                                    Citing \\ Answers
                                                        [
                                                            \cite{GopherCite-Menick2022TeachingLM}{,}
                                                            \cite{WebGPT-Nakano2021WebGPTBQ}
                                                            , leaf, text width=25em
                                                        ]
                                                ]
                                        ]
                                ]
                                [
                                    Non-RL \\ Training
                                        [
                                            \cite{RAFT-Dong2023RAFTRR}{,}
                                            \cite{Yuan2023RRHFRR}{,}
                                            \cite{Scheurer2022TrainingLM}
                                            , leaf, text width=25em
                                        ]
                                ]
                        ]
                        [
                            Reward \\ Alternatives \\ (\S \ref{sec:alternatives})
                            [
                                    \cite{SLiC-HF-Zhao2023SLiCHFSL}{,}
                                    \cite{Liu2023ChainOH}{,}
                                    \cite{Bai2022ConstitutionalAH}{,}
                                    , leaf, text width=25em
                                ]
                        ]
                ]
        \end{forest}
    }
    \caption{Categorization of different components in the RLHF and example representative works from literature. 
    }
    \label{fig:categorization_of_reasoning_big}
\end{figure*}

%% file: LaTeX/survey_sections/200_rlhf_overview2.tex
\begin{figure}[t]
    \centering
    \includegraphics[width=0.99\columnwidth]{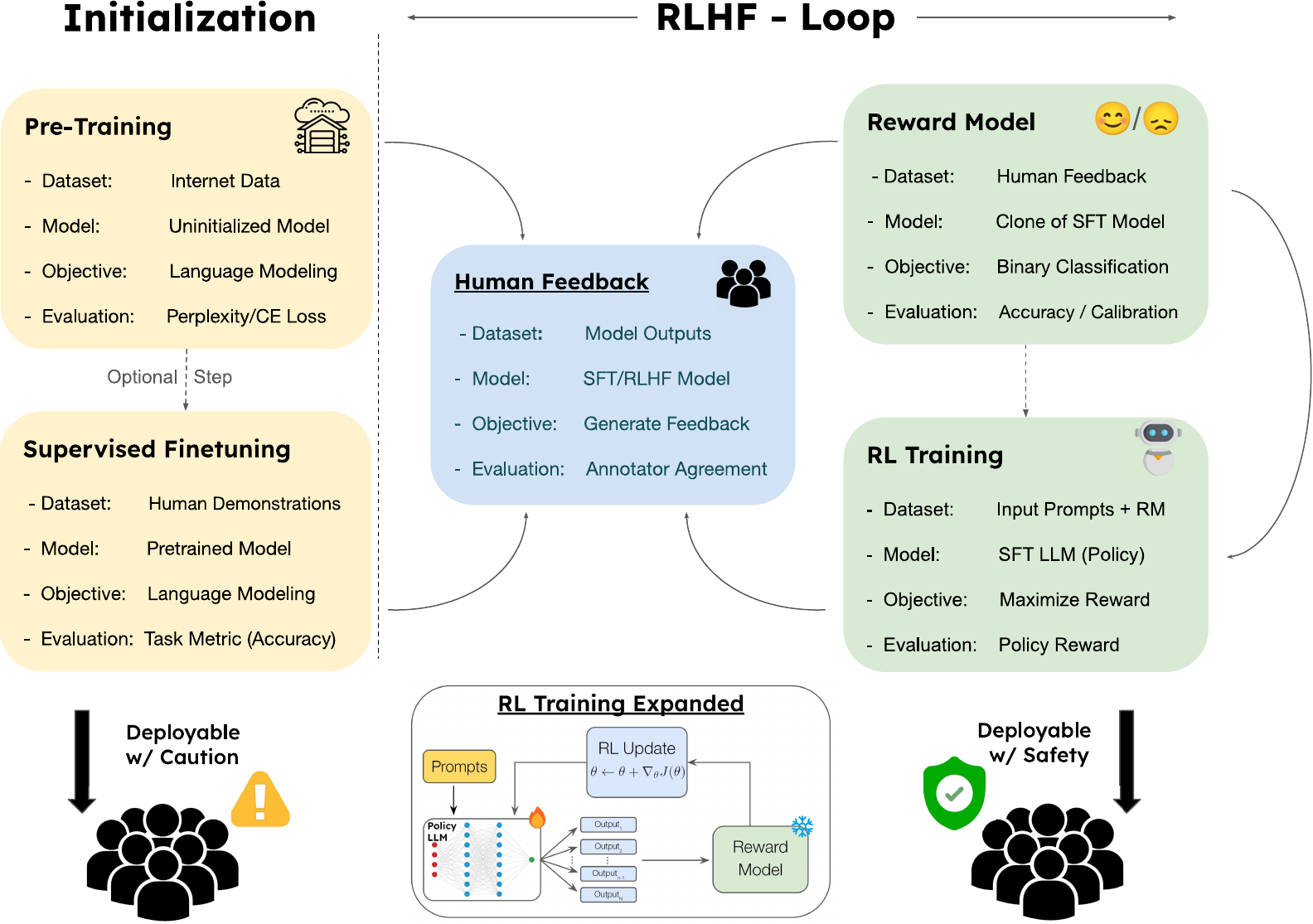}
    \caption{Workflow of RLHF. A pretraining phase, and optionally supervised finetuning (SFT) on human demonstrations, is followed by all RLHF workflows for training language models. This is followed by an iterative loop starting with collecting human feedback on model-generated outputs, training a reward model, and updating the language model using a suitable RL algorithm.}
    \label{fig:rlhf_overview}
\end{figure}

\subsection{Reinforcement Learning from Human Feedback (RLHF): Overview and Motivation}

\paragraph{The Importance of Human Feedback in Language Models}

The alignment of a model with the user's intentions and preferences is critical, and incorporating human feedback in model training is a key step towards achieving this (Section \ref{sec:motivation}). However, the process of obtaining high-quality human feedback, particularly in the form of human demonstrations, can be a resource-intensive process, both in terms of time and cost. A more efficient approach is to collect feedback on the outputs generated by the model and train the language model to incorporate this feedback. However, collecting such a large amount of feedback is also costly and impractical for real-time/online collection during training.

\paragraph{The Role of RLHF in Language Models}

Reinforcement Learning from Human Feedback (RLHF) offers a solution to these challenges. In RLHF, human feedback is collected offline and used to train a reward model. This reward model then acts as a surrogate for human feedback during training, providing reward signals to the Language Model.
Reinforcement learning algorithms form the natural candidates for training a model from scalar evaluative feedback, as provided by the reward model.
This forms the essence of  Reinforcement Learning from Human Feedback (RLHF)~\cite{DeepRLHF-Christiano2017DeepRL} as used to train Language Models. This approach is more sample-efficient and has shown more promising results compared to supervised fine-tuning alone~\cite{InstructGPT-Ouyang2022TrainingLM}.

\paragraph{Applications of RLHF in Language Models}
In early works, RL has been used in training Language models across various domains such as dialogue generation~\cite{Yi2019TowardsCA,DialogueHITL-Li2016DialogueLW,WayOff-Jaques2019WayOB}, machine translation~\cite{Kreutzer2018CanNM, Nguyen2017ReinforcementLF,QualityAware-Fernandes2022QualityAwareDF, Sokolov2016BanditSP}, text generation ~\cite{TextGen-Li2017ParaphraseGW,TextGen-Shi2018TowardDT,TextGen-Zhou2020LearningTC,TextGen-Ziebart2008MaximumEI}, semantic parsing~\cite{Lawrence2018ImprovingAN}, summarization~\cite{SummarizeHF-Stiennon2020LearningTS, FinetuningRLHF-Ziegler2019FineTuningLM, RecursiveSumm-Wu2021RecursivelySB}.
More commonly, these methods were trained using non-differentiable automated evaluation metrics such as BLEU, ROUGE~\cite{Rogue-Ranzato2015SequenceLT, Rogue-Shen2015MinimumRT, Rogue-Keneshloo2018DeepRL}, or simulated feedback~\cite{Nguyen2017ReinforcementLF}.
However, while the combination of RL and human feedback has been extensively studied \cite{Knox2008TAMERTA,DeepRLHF-Christiano2017DeepRL}, it is only recently that RLHF with LLMs has achieved significant success in sequence-to-sequence tasks such as Summarization~\cite{SummarizeHF-Stiennon2020LearningTS,FinetuningRLHF-Ziegler2019FineTuningLM, RecursiveSumm-Wu2021RecursivelySB}, providing reliable answers with citations to queries~\cite{WebGPT-Nakano2021WebGPTBQ, Sparrow-Glaese2022ImprovingAO}, creating Helpful, Harmless and Honest dialogue agents aligned with broad human values~\cite{InstructGPT-Ouyang2022TrainingLM,HHHAssistant-Bai2022TrainingAH}.

\paragraph{Formulating Language Modeling as a RL Problem}
Reinforcement Learning (RL) is a learning paradigm
for a setting where an agent must make a sequence of decisions while interacting with an environment and obtaining evaluative feedback in the form of rewards. The agent's objective is to maximize the total reward it receives over time. In the context of language models, the agent is the language model itself, and its actions consist of generating tokens from its vocabulary. The agent's policy, which maps states to actions, is represented by the language model's parameters. The agent receives rewards from the environment, which in this case is a reward function that forms a surrogate from human feedback (Section~\ref{sec:role_of_reward}). The agent's objective is to optimize its actions (by updating its policy) to maximize the cumulative reward. A thorough mathematical formulation can be found Section~\ref{sec:llm_rl_formulation}, and has been summarized in
Table~\ref{tab:rl_components} and Figure \ref{fig:output_mdp}. While these details are sufficient for further discussion in the paper, we refer interested readers to \citet{Arulkumaran2017DeepRL,sutton2018reinforcement} for more details about reinforcement learning.

\begin{table}[!t]
    \centering
    \centering
    \begin{tabular}{ll}
        \toprule
        \textbf{RL Component} & \textbf{Language Model Equivalent}        \\
        \midrule
        Agent                 & Language Model                            \\
        State                 & Input prompt + currently generated text   \\
        Action                & Predicted next token from vocabulary      \\
        Policy                & Output Distribution of language model     \\
        Reward                & Evaluative Feedback from the reward model \\
        Trajectory            & Input + completely generated text         \\
        \bottomrule
    \end{tabular}
    \captionsetup{skip=7pt}
    \caption{Mapping of terms used in RL literature to training of language models through RLHF. See Table~\ref{tbl:formulation} for mathematical formulation.}
    \label{tab:rl_components}
\end{table}

\paragraph{The Workflow of RLHF}

RLHF, as first popularized by \cite{DeepRLHF-Christiano2017DeepRL} for mastering Atari Games consists of three crucial stages.
An overview of standard RLHF workflow is highlighted in Figure~\ref{fig:rlhf_overview}.
The first stage involves the collection of human feedback on a set of \texttt{<input, output>} pairs. These pairs can be sourced from existing datasets or generated by the pre-trained model for a given set of input prompts. The second stage involves learning a reward model from the collected human feedback. The reward model is trained to output a scalar reward for a given \texttt{<input, output>} pair, indicating the favorability of the pair. In essence, the reward model is trained to mimic human feedback, such that for a given input, desirable outputs are scored higher than undesirable outputs.
The final stage involves the RLHF training of the language model, where the reward model provides reward signals on model outputs, usually in the form of scalar reward. The parameters of the language model are then updated based on these reward signals using an appropriate policy-gradient RL algorithm, updating the model to produce more rewarding outputs.

These stages can be performed iteratively, with the intermediately trained model generating more prompts to collect additional human feedback. This feedback is then used to train the reward model, and the process is repeated multiple times~\cite{SummarizeHF-Stiennon2020LearningTS, HHHAssistant-Bai2022TrainingAH, GopherCite-Menick2022TeachingLM}.
In the following sections, we discuss each of these stages in detail. We start with Human Feedback Collection (Section~\ref{sec:human_feedback}), followed by training the Initial Policy (Section~\ref{sec:inital_policy}), Reward Model Training (Section~\ref{sec:reward_model}), and finally RLHF Training (Section~\ref{sec:rlhf_training}).
Finally, we discuss the properties of RLHF-trained models and their limitations in Section~\ref{sec:rlhf_properties}.

%% file: LaTeX/survey_sections/210_human_feedback.tex
\subsection{Human Feedback}
\label{sec:human_feedback}

In this section, we discuss the nature, objectives, and different types of human feedback, followed by the challenges and strategies associated with collecting high-quality feedback.

\subsubsection{Nature and Objectives of Human Feedback}

Tasks such as summarization and providing helpful answers are inherently ambiguous and require human judgment to evaluate the quality of the generated text.
Automated metrics like BLEU and ROUGE~\cite{Rogue-Lin2004AutomaticEO} often do not correlate with human judgment~\cite{BadEval-Liu2016HowNT, BadEval-Schluter2017TheLO,Sellam2020BLEURTLR,SummarizeHF-Stiennon2020LearningTS}, making them unreliable for evaluation and training.
Thus, acquiring high-quality human feedback to align the model with human behavior becomes crucial.
Feedback is typically provided on the outputs generated by the model (or input-output pairs from the dataset), and subsequently, the model is trained to learn from this feedback.
However, capturing diverse human preferences is a challenging task.
One approach to encapsulate subjective human preferences is to approximate them using ``models of human behavior''.
This concept of human behavior models has roots in diverse fields such as econometrics~\cite{McFadden1981EconometricMO}, psychology~\cite{OConnor1989ModelsOH}, and inverse reinforcement learning.
A notable example is the Bradley-Terry model~\cite{Bradley1952RankAO}, a probabilistic model that encodes the preference of one output over another in pairwise competitions. In the context of RLHF, reward models that form surrogates for human preferences serve as such models of human behavior.

The type of feedback collected depends on the intended objective to be displayed by the fine-tuned language model.
\citet{AlignLab-Askell2021AGL} proposes three objectives for an aligned Language Model: Helpfulness, Honesty, and Harmlessness (HHH). These objectives can be broadly defined as follows:

- \textbf{Helpful:} A Language Model is considered helpful if it can efficiently complete tasks or answer questions (while being harmless), ask relevant follow-up questions when necessary, and appropriately redirect ill-informed requests. Helpfulness includes context-dependent aspects such as informativeness, coherence, relevance, creativity, and specificity.

- \textbf{Honest:} Honesty in a Language Model implies providing accurate information, expressing appropriate levels of uncertainty, and honestly conveying its capabilities, knowledge, and internal state.
Language Models are particularly susceptible to hallucination~\cite{Khandelwal2019SampleET, Maynez2020OnFA}, making it essential to penalize such behavior. Unlike helpfulness, honesty is more objectively evaluated.

- \textbf{Harmless:} A harmless Language Model should avoid offensive or biased behavior, refuse to aid in dangerous acts, recognize disguised nefarious attempts, and act with modesty and care when providing advice with potentially sensitive or consequential impacts.

\future{Make sure the next statement is clear to audience.}
These broad objectives, as mentioned above, encompass specific objectives, which can be considered subcategories.
For example, in the case of summarization, the summary should be helpful to the reader and should not contain any false or harmful information.
Similarly, the goal of reducing bias in a dialogue agent's responses can be considered a subset of the Harmless objective. At the same time, coherence and creativity in the generated text are aspects of being helpful.
These objectives are not mutually exclusive and are context and task-dependent. Even human labelers and researchers have shown disagreements in annotation~\cite{Kreutzer2018ReliabilityAL}.

\subsubsection{Types of Human Feedback}

Human Feedback is usually collected on model-generated outputs. Good feedback should incorporate information on where the model output is lacking and how to improve it. A simple process is to let human labelers provide feedback on a set of model outputs generated from a dataset of prompts or inputs.
Alternatively, existing datasets can be repurposed to incorporate implicit feedback, such as rating different user choices~\cite{Kreutzer2018CanNM}.
Regardless of the process, human feedback can be collected in various forms, such as binary responses, preference ranking, language feedback, etc. While the choice of feedback type depends on the downstream task, it is essential to note that the feedback should be collected in a way that is easy for humans (labelers) to provide; there is high agreement among the labelers, and it is also informative.
In this section, we classify the feedback into four different categories: rating feedback, ranking feedback, language feedback, and miscellaneous feedback. 

\paragraph{Rating Feedback}

The simplest form of rating feedback is binary feedback, where the labeler is asked to provide a binary response (yes/no) to a given input~\cite{DialogueHITL-Li2016DialogueLW,ScaleLanguageFeedback-Scheurer2023TrainingLM}.
Binary feedback is easy to collect and interpret. Some works have used binary responses to get feedback on multiple questions (such as if the generated text is coherent)~\cite{Yi2019TowardsCA}.
A richer form of feedback is to ask labelers to provide a rating on a scale.
The scale can be continuous~\cite{Graham2013ContinuousMS}, or be similar to Likert Scale~\cite{likert1932technique} (where user rate using an integer from 1 to k)~\cite{Kreutzer2018CanNM, WayOff-Jaques2019WayOB}. A different variant of rating feedback is to provide categorical feedback such as `incorrect', `partially-correct', and `correct'~\cite{Gao2023ContinuallyIE}.
While rating feedback is easy to specify, often inter-annotator agreement is low because of the subjective nature of the task~\cite{Kreutzer2018ReliabilityAL}. Further, the order of examples presented to the annotator may bias the results~\cite{Yannakakis2011RankingVP}.
Moreover, it is challenging to differentiate between data points with outputs of similar quality since feedback is provided individually to each output without comparison.

\paragraph{Ranking or Preference Feedback}

Ranking feedback or Preference-based feedback has been extensively used in the recent development of AI assistants and found to be both convenient to collect and performative. Specifically, the labeler is offered with binary \cite{SummarizeHF-Stiennon2020LearningTS} or multiple choice options~\cite{FinetuningRLHF-Ziegler2019FineTuningLM}, and asked to select the most appropriate response based on a certain set of instructions (directions).
Recently, \cite{Zhu2023PrincipledRL} has shown convergence guarantees for reward models trained using this feedback form.
Moreover, given an input prompt, it is common to ask labelers to rank k (> 2) generated responses, which are then repurposed as pairwise comparisons for the reward model~\cite{InstructGPT-Ouyang2022TrainingLM}.
However, collecting pairwise feedback might still be difficult for near similar responses and may result in much time spent by the labelers even on single input~\cite{ScaleLanguageFeedback-Scheurer2023TrainingLM}.
Additionally, preference-based feedback provides a very sparse signal, conveying limited information about the reasoning behind the provided feedback. Moreover, it is provided only on the complete text generated by the model (trajectory) and not on specific parts of the text (particular state)~\cite{Pang2022RewardGI, lewis-etal-2017-deal}.
Moreover, preference-based feedback provides no further improvement in terms of inter-annotator agreement when compared to rating feedback~\cite{Kreutzer2018ReliabilityAL}.

\paragraph{Language Feedback}

A more informative way to provide feedback is in free-form language. This provides a dense reward signal, specifying more precisely where the model goes wrong or needs improvement.
For example, consider the case where the output generated by the model is ``A humorous story about a specific profession involving person A and person B.`` The previous feedback forms would provide only sparse signals, such as indicating that the output is inappropriate. However, this feedback alone will not help the model identify the cause of inappropriateness, and the single example alone can imply that the text is inappropriate because: ``it is wrong to create humor in general,`` ``it is wrong to create humor about specific professions`` or ``it is wrong to involve individuals in humorous stories`` and so on. On the other hand, free-form feedback can provide more precise feedback, such as ``It is inappropriate to create humor that targets specific professions.`` This enables the model to understand the issue from a single example better and generalize to similar cases without learning from more examples.

Language Feedback has been extensively used in various domains such as Dialogue models~\cite{DialogueHITL-Li2016DialogueLW, FeedYourself-Hancock2019LearningFD}, Summarization~\cite{ScaleLanguageFeedback-Scheurer2023TrainingLM}, Question-Answering~\cite{Li2022UsingIF}, Code generation~\cite{Chen2023ImprovingCG}. Recently, \cite{ScaleLanguageFeedback-Scheurer2023TrainingLM} has shown that language feedback is more effective than preference-based feedback in the context of summarization systems.
Also, as \cite{FeedYourself-Hancock2019LearningFD} discusses, getting preference-based feedback is plausible for paid labelers but not for real users using real deployed systems. Real users interact with the system through free-form language; hence, getting human feedback in the free-form language is more natural.
Although task-dependent, \cite{ScaleLanguageFeedback-Scheurer2023TrainingLM} further find that labelers take only 3x times to provide language feedback compared to preference-based feedback, despite providing much granular information.
However, incorporating language feedback in the RLHF pipeline is not straightforward, and there has been limited work in this direction.

\paragraph{Miscellaneous Feedback}

Apart from providing single feedback, methods have experimented with using a combination of feedback types or altogether different types. For example, \cite{Sparrow-Glaese2022ImprovingAO} uses a combination of rule violation feedback (binary), preference-based feedback, and rating of evidence.
\cite{DeepMindSolvingMath-Uesato2022SolvingMW,PretrainingRLHF-Korbak2023PretrainingLM} provide segment-level feedback instead of the whole text, and \cite{FineGrainedRLHF-Wu2023FineGrainedHF} provide feedback at the token level. Moreover, some studies employ indirect methods for collecting feedback. For example, \cite{Kreutzer2018CanNM} uses human interactions on translated eBay titles to find more preferred translations.

Further, it is also possible to provide computational feedback, for example, from automated metrics~\cite{A2C-Bahdanau2016AnAA}, forms of synthetic feedback~\cite{Kim2023AligningLL, Black2023TrainingDM}, web descriptions~\cite{Hanjie2022SemSupSS, Aggarwal2023SemSupXCSS}, LLM generated feeedback~\cite{Shinn2023ReflexionAA, Madaan2023SelfRefineIR, Yang2022Re3GL},  which might, in turn, be generated based on certain human requisites or instructions~\cite{Bai2022ConstitutionalAH, Sun2023PrincipleDrivenSO, kundu2023specific}. However, these methods still use little to no human feedback and may have several unexplored limitations such as instability and lack of robustness~\cite{shumailov2023curse, alemohammad2023selfconsuming, gudibande2023false} and are not the focus of this survey. We refer readers to \citet{Fernandes2023BridgingTG} for discussion on different type of feedback used in Natural Language Generation.

\subsubsection{Collection of High-Quality Human Feedback}

Collecting high-quality human feedback is a challenging task that has been the focus of extensive research.
The quality of feedback is pivotal; subpar or noisy feedback can significantly hamper the performance of the final trained model.
For example, for summarization tasks, \cite{FinetuningRLHF-Ziegler2019FineTuningLM} discovered that their model predominantly extracted verbatim lines from the document. This was later attributed to low-quality feedback by \cite{SummarizeHF-Stiennon2020LearningTS}.
Similarly, the size of the feedback is also crucial.
For example, despite employing similar methodologies, \cite{LetsVerify-Lightman2023LetsVS} identified a need for a `greater amount of feedback' for the methods in \cite{DeepMindSolvingMath-Uesato2022SolvingMW} to be effective, as the intended objective was not even observed in the latter work.

The provision of clear and unambiguous instructions to the labelers is a fundamental requirement~\cite{FinetuningRLHF-Ziegler2019FineTuningLM,WebGPT-Nakano2021WebGPTBQ}.
Failure to do so can not only result in low-quality feedback but also introduce systematic bias in the collected feedback and, consequently, the model~\cite{Parmar2022DontBT}.
Typically, labelers are provided with a comprehensive set of instructions, including guidelines for handling edge cases~\cite{HHHAssistant-Bai2022TrainingAH}. \cite{Sparrow-Glaese2022ImprovingAO} even provides a tutorial to the selected few labelers.

Researchers typically screen labelers to ensure they possess the necessary skills to provide feedback. For instance, in the case of translation tasks, bilingual labelers with native proficiency in both languages are preferred~\cite{Kreutzer2018ReliabilityAL}.
Additionally, a minimum educational qualification is generally preferred.
For example, \cite{SummarizeHF-Stiennon2020LearningTS} requires labelers to have at least a high-school degree, whereas \cite{WebGPT-Nakano2021WebGPTBQ}, \cite{Sparrow-Glaese2022ImprovingAO} and \cite{HHHAssistant-Bai2022TrainingAH} require a minimum undergraduate and master's degree respectively.
The end goal also influences the selection of labelers. For instance, creating a harmless and helpful chatbot necessitates a diverse group of labelers with varying backgrounds and demographics~\cite{InstructGPT-Ouyang2022TrainingLM,HHHAssistant-Bai2022TrainingAH} as otherwise this may result in implicit biases in the model~\cite{peng2022investigations}. For instance, currently deployed language models have been shown to reflect views more aligned with western audiences~\cite{Durmus2023TowardsMT} and may have systematic political biases~\cite{santurkar2023opinions}, partly owing to the lack of annotators from diverse demographic groups.

However, despite screening, there may be low agreement among the annotators themselves, or even between researchers and annotators~\cite{Kreutzer2018CanNM}.
The labelers are further screened based on two standard criteria 1.) inter-annotator agreement, i.e., the agreement between different annotators on the same example, and 2.) expert-annotator agreement, i.e., the agreement between annotators and experts \cite{Kreutzer2018ReliabilityAL}.
Specifically, the former metric ensures that the labelers are consistent in their feedback, and the latter metric is used to keep only those labelers that have a high agreement with experts. \cite{GopherCite-Menick2022TeachingLM} creates a group of super-raters who have a high agreement with experts, and the group is expanded upon iteratively.
Even after filtering, some methods ensure a hands-on relationship with labellers~\cite{SummarizeHF-Stiennon2020LearningTS} and have also created Slack groups for discussing any bugs, issues, or edge cases~\cite{HHHAssistant-Bai2022TrainingAH}.

%% file: LaTeX/survey_sections/220_initial_policy.tex
\subsection{Supervised Fine-Tuning: Limitations and Role}
\label{sec:inital_policy}

Upon the collection of high-quality feedback, the subsequent step is to assimilate this feedback to train the model. The most direct method to achieve this is to perform supervised fine-tuning of the language model based on the collected feedback. Specifically, human feedback is gathered in the form of expert outputs on input prompts, also referred to as human demonstrations. These human demonstrations can be perceived as positive example outputs to prompts that should be generated by the language model. The model is then fine-tuned on these demonstrations using the same pretraining objective, and this process in RL terminology is often termed behavior cloning~\cite{WebGPT-Nakano2021WebGPTBQ}.

Additionally, when dealing with preference data, the model can be directly fine-tuned on preferred feedback. However, this approach exhibits limitations by not accounting for negative feedback—outputs that the model should avoid generating. This is crucial for training robust models that can handle adversarial situations and identify and rectify errors. To tackle this limitation, alternative methods that incorporate both positive and negative feedback have been developed, as discussed in Section~\ref{sec:alternatives}.

In addition to human demonstrations, existing public instances from NLP datasets can be used as instruction tuning demonstrations~\cite{Wei2021FinetunedLM}.
This usually involves creating new instruction-tuning datasets by adding task instructions to existing examples from the dataset~\cite{Ajith2023InstructEvalSE}.
In another field of work, prompts from the initial iterations of GPT-3~\cite{Brown2020LanguageMA} served to real customers through Web API were used to fine-tune the model on expert (human) demonstrations provided by contracted labelers~\cite{InstructGPT-Ouyang2022TrainingLM}.

\paragraph{Limitations of Supervised Finetuning}

While finetuning on supervised data enhances the model beyond its pretrained version in following instructions and intended tasks, it suffers from numerous limitations.
For instance, it does not penalize the model for hallucinating or permit it to learn from neutral or negative feedback. This can lead to harmful and unintended behavior, making it easier to prompt such models to elicit them~\cite{Ganguli2022RedTL, RedTeaming-Perez2022RedTL}.
Furthermore, behavior cloning is likely to perform poorly in out-of-distribution prompts~\cite{pomerleau1988alvinn}. These limitations may stem from the fact that during behavior cloning, the model is not allowed to \textit{explore} the vast space of possible actions, i.e., the model is not allowed to generate outputs that are not present in the demonstrations and, in turn, get feedback for them. We refer readers to Section~\ref{sec:why_rl} for theoretical discussion on the limitations of SFT.

\paragraph{SFT as Initial Policy in RLHF models}

Despite its caveats, supervised fine-tuning plays a pivotal role in RLHF as it provides a robust initial policy, which allows RLHF methods to work well.
From an RL perspective, learning algorithms such as the widely used Proximal Policy Optimization (PPO) in training sequence-to-sequence models, struggle to improve from poor initializations, especially when the action space is large, as in the case of text generation. This is because these methods use model-based exploration, which is ineffective when the transition probabilities over many actions are similar~\cite{Nguyen2017ReinforcementLF} i.e., different text outputs have similar probabilities of generation.

Furthermore, as we discuss in Section~\ref{sec:rlhf_training}, usually a KL penalty is applied to ensure the output text generated by our RL-tuned model is close to the initial model. Thus, during RL training, it is preferable to start with an initial model that already generates decent-quality text.

Empirical studies have demonstrated that starting with fine-tuning on high-quality human demonstrations results in significant improvements over starting with pretrained language models~\cite{SummarizeHF-Stiennon2020LearningTS,InstructGPT-Ouyang2022TrainingLM}. For instance, InstructGPT collects API customer and labeler written prompts and outputs to fine-tune their model before initiating with the RLHF training~\cite{InstructGPT-Ouyang2022TrainingLM}.

\cite{Sparrow-Glaese2022ImprovingAO} has also shown that starting with a prompted model (dialogue) instead of fine-tuning on label demonstrations is possible. However, they start with a large model (70B parameters) and do not perform a comparative study starting with fine-tuning on human demonstrations. Thus, it cannot be definitively concluded that starting with a prompted model is equivalent to fine-tuning on human demonstrations. Moreover, prompting has the limitation of using up a major portion of the context length of the model, which, apart from the computational burden, can also be crucial for some tasks because of limited context length. \cite{AlignLab-Askell2021AGL} propose using context distillation by training the model to generate output similar to its prompted counterpart using KL divergence loss. They find similar performance to the prompted model, and the method has been used in their subsequent works~\cite{HHHAssistant-Bai2022TrainingAH}.

In conclusion, while supervised fine-tuning can be utilized independently of the RLHF pipeline, it still suffers from several significant limitations. However, it still serves as an integral step in the RLHF pipeline, providing a robust initial policy crucial for subsequent RL training.

%% file: LaTeX/survey_sections/230_reward_modelling.tex
\subsection{Reward Modeling}
\label{sec:reward_model}

\paragraph{Reward as a Proxy for Human Feedback}
After the collection of human feedback, the next challenge is training the language model effectively. Although supervised fine-tuning offers a straightforward method, its effectiveness is limited by the volume of human feedback.
In contrast, RLHF introduces a reward model to emulate human feedback, thereby acting as a stand-in for the true reward function, i.e., the actual human feedback. This reward model, usually much smaller than the language model, facilitates fine-tuning the language model using feedback generated by it on new model outputs, avoiding the need for additional costly human annotation. In practice, using a reward model over supervised fine-tuning has been found more data-efficient~\cite{Ramamurthy2022IsRL}.

\paragraph{Training a Reward Model}

The reward model is a fine-tuned language model that assigns a scalar reward score to an input-output pair. The last embedding layer is replaced with a single projection layer that outputs this scalar reward.
While the reward model can learn from various types of feedback, recent studies highlight the simplicity and effectiveness of preference-based feedback~\cite{InstructGPT-Ouyang2022TrainingLM,HHHAssistant-Bai2022TrainingAH}.
This approach involves fine-tuning the initialized reward model to predict the preference between two trajectories (output text) given the same input prompt or context. The reward is typically modeled as a Bradley-Terry-Luce (BTL) model~\cite{Bradley1952RankAO}, where the probability of preferring one trajectory over another is a function of the difference in their reward scores.
Mathematically, this can be represented as:
\begin{align}
    \Pr((o \succ o', c) \mid \phi) = \sigma[R_\phi(c, o) - R_\phi(c, o')]
\end{align} where $\sigma(x) = \frac{1}{1 + e^{-x}}$ is the sigmoid function, o and o' represent the two trajectories, and their rewards are represented as $R_\phi(c, o)$ and $R_\phi(c, o')$ respectively.
This form of reward modeling has been found to provide smoother rewards and is less noisy~\cite{DeepRLHF-Christiano2017DeepRL}. A similar method can then be used for ranking between k trajectories (k > 2), where the reward is modeled as a Plackett-Luce (PL) model~\cite{Plackett1975TheAO,Luce1979IndividualCB}. Moreover, \cite{Zhu2023PrincipledRL} provides theoretical proof of convergence guarantees under the Maximum Likelihood estimate of both BTL and PL models.

The size and initialization of the reward model are critical determinants of its performance.
While smaller reward models are easier to train, scaling laws suggest that larger models yield better agreement with actual human preferences~\cite{AlignLab-Askell2021AGL}.
However, \cite{InstructGPT-Ouyang2022TrainingLM} found that training very large reward models can be unstable and result in overfitting. Instead, they report good performance even when using a reward model that is 30 times smaller than the policy model.

Regarding initialization, multiple methods have been proposed. While \cite{FinetuningRLHF-Ziegler2019FineTuningLM} fine-tunes a pretrained language model on preference data collected on model-generated outputs, \cite{InstructGPT-Ouyang2022TrainingLM} trains a GPT-3 based reward model on publicly available datasets. However, only a slight advantage was found over using pretrained language models or supervised-fine-tuned models.
Leveraging publicly available preference datasets (such as ranked answers from StackOverflow), as suggested by \cite{AlignLab-Askell2021AGL}, notably enhances reward model performance, especially for smaller models and datasets.

\paragraph{Challenges in Reward Modeling}

The reward model is initially trained on a selected set of input prompts and corresponding initial model outputs. As the model training progresses, it is crucial for the reward model to generalize to new model outputs and potentially new input prompts. We refer readers to Section~\ref{sec:analysis_reward_model} for a deeper theoretical exploration of this aspect.

Regarding the generalization capabilities of reward models, \citet{InstructGPT-Ouyang2022TrainingLM} presents findings that demonstrate high generalization to held-out test labelers. This capability is of paramount importance since a majority of the inputs encountered during language model training would be out-of-distribution w.r.t. the reward model training phase. Generalization capability depends on various factors such as the dataset's size, the amount of noise in the feedback dataset, and the characteristics of the pretrained reward model.

Moreover, the robustness and calibration of the reward models with respect to actual human preferences are essential for their effectiveness. A well-calibrated reward model should accurately predict the probability of a human preferring one output over another.
\citet{HHHAssistant-Bai2022TrainingAH} discovered that when training solely on a helpfulness feedback dataset, their model exhibits strong calibration. However, when trained on a mixture of helpfulness and harmlessness datasets, the model is underconfident in its predictions. To assess robustness, a common practice involves evaluating the policy model trained using the reward model.

To assess robustness, a common practice involves evaluating the policy model trained using the reward model.
Interestingly, \citet{HHHAssistant-Bai2022TrainingAH} discerned that smaller reward models and higher rewards correlate with decreased robustness. This phenomenon arises from the reward model's initial training on model outputs with naturally low rewards. To address this distribution shift, an approach involving iterated training of the reward model is proposed (see Section~\ref{sec:rlhf_training}). In summation, the discussion underscores that the trained reward model on preferences is an imperfect proxy of human feedback, especially in out-of-domain cases.

\paragraph{Moving Beyond Scalar Rewards}

Apart from providing a single scalar reward at the end of a trajectory (complete text output), several methods model a more fine-grained approach. \cite{DeepMindSolvingMath-Uesato2022SolvingMW,PretrainingRLHF-Korbak2023PretrainingLM} provides a segment-level reward during training, a method also known as process supervision. Interestingly, while \cite{DeepMindSolvingMath-Uesato2022SolvingMW} did not find any major downstream performance improvement with their method, \cite{LetsVerify-Lightman2023LetsVS} used similar methodology but instead trained larger models on a larger feedback dataset coupled with evaluation on a more difficult task found segment-level feedback to be significantly more useful. \cite{ScaleLanguageFeedback-Scheurer2023TrainingLM} uses language feedback from another LLM that implicitly acts like a reward model for the training of the policy model.

While ideally, as discussed in Section~\ref{sec:role_of_reward}, the reward model provides a dual-purpose reward taking into account both the task information (eg, summarization task) and the task-specific evaluation ((a condescending summary is rewarded less than a neutral summary). However, diversifying the approach, some strategies involve the use of multiple reward models, each specializing in distinct characteristics or specific tasks. \cite{FineGrainedRLHF-Wu2023FineGrainedHF,RewardSoup-Ram2023RewardedST} demonstrate the efficacy of training separate reward models for specific attributes such as coherency and factuality. Similarly, \cite{Sparrow-Glaese2022ImprovingAO} introduces two reward models—one for preference and another for rule violation in dialogue generation. They found using two models over one to be more effective, likely because of a smaller feedback dataset.
Further, since the preference-based reward model provides a delayed reward (reward is provided only at the end of the whole trajectory), the A2C algorithm, when used for sequence modeling \cite{A2C-Bahdanau2016AnAA} proposes potential-based reward shaping, where intermediate generations (states) are also rewarded.

In conclusion, the reward modeling process is a critical component of RLHF which involves the training of a model to emulate human feedback, thereby acting as a surrogate for the true reward function. The size, initialization, and generalization capabilities of the reward model are all crucial factors that influence its performance. The reward model must be robust, well-calibrated, and additionally can provide more fine-grained feedback to the policy model training.

%% file: LaTeX/survey_sections/240_finetuning.tex
\subsection{RLHF Finetuning of Language Models}
\label{sec:rlhf_training}

The trained reward model is utilized for finetuning the language model. Framing the task as reinforcement learning, with the language model as the policy, algorithms such as Proximal Policy Optimization (PPO) and Advantage Actor-Critic (A2C) ~\cite{PPO-Schulman2017ProximalPO, A2C-Bahdanau2016AnAA} are used to update the parameters of the language model such that the generated outputs maximize the obtained reward.
These are gradient-based methods, called policy-gradient algorithms, that directly update the parameters of the policy using the evaluative reward feedback
The following sections primarily focus on the widely used Proximal Policy Optimization (PPO) Algorithm, while the same concepts are applicable to other candidate algorithms~\cite{InstructGPT-Ouyang2022TrainingLM}.

\subsubsection{Training Procedure}

The pre-trained/SFT language model is prompted with contexts/prompts from a prompting dataset. The prompting dataset may or may not be identical to the one used for collecting human demonstrations in the SFT phase~\cite{InstructGPT-Ouyang2022TrainingLM}. The model outputs, along with the inputs, are passed to the reward model that generates a scalar output indicating the reward for this input-output pair. The reward is used as evaluative feedback to update the parameters of the language model using suitable RL algorithms that result in increasing the likelihood of the generation of more rewarding outputs. We next discuss a few commonly used RL algorithms for the process.

\subsubsection{Training Algorithms}
The commonly used policy-gradient algorithms for aligning LLMs using RLHF are PPO and A2C~\cite{PPO-Schulman2017ProximalPO, A2C-Bahdanau2016AnAA}. Both fall under the category of actor-critic algorithms.
These algorithms consist of two main components: the critic learns the expected cumulative reward for an input-output pair, called the value function, and the actor is the LLM policy that gets updated based on the cumulative reward estimates obtained from the critic. The reward values are obtained from the previously trained reward model, which is kept frozen during the RL training.
As the LLM encounters more interactions and collects more reward feedback, it uses the data to update the value function and the LLM policy parameters.
The training objective (Equation \eqref{eq:gen_pg}) aims to update the parameters of the policy to increase the expected cumulative reward of the LLM policy.
A2C and A3C \cite{mnih2016asynchronous} use an estimate of the \textit{advantage} of taking an action instead of the action-value function for that action as a way of incurring lesser variance in policy gradient estimation.
PPO additionally constrains the policy update at each iteration from straying too far by using a clipped objective \cite{PPO-Schulman2017ProximalPO}. This helps provide additional stability to the training.
Training LLMs at a large scale requires an immense engineering effort, and practical implementations of these algorithms require domain-specific variations.
While major progress has been made towards efficient training and inference of LLMs~\cite{xiasheared,lagunas2021block, cofi, supru, murahari2023mux, yang-etal-2022-textpruner,Hinton2015DistillingTK, yin2021autotinybert, datamuxmurahari}, there is still a lot of scope for improvement in the sample efficiency and stability of training algorithms for RLHF.
Recent work has addressed these challenges with different variants of these algorithms tackling different aspects ranging from practical implementation issues such as high memory usage \cite{santacroce2023efficient}, changes specific for NLP \cite{Ramamurthy2022IsRL, Wu2023PairwisePP}, training instability~\cite{Zheng2023SecretsOR}.
We refer readers to \cite{weng2018PG} for a comprehensive survey of policy gradient algorithms.

\subsubsection{Improving Training Stability}

Imperfections in the reward model reduce the effectivity of the training algorithms, as the value functions learned, and in turn thus the gradient updates, become inaccurate.
Thus, using the aforementioned algorithms with the learned reward model may lead the language model to exploit the imperfections and generate nonsensical text, often called `reward overoptimization'.
This can be mitigated with appropriate regularization during training.
As the pre-trained or SFT model (policy) is already a highly capable LLM, \citet{WayOff-Jaques2019WayOB} propose using a copy of the initial model to regularize training.
The aim is to ensure that even as the policy parameters are updated to maximize reward, the outputs of the updated policy do not stray too far from the initial policy.
In particular, an additional regularization term of the Kullback-Leibler (KL) divergence between the policy being trained and the initial policy is added to the RL training objective in the form of a reward penalty, commonly called the KL penalty.
Theoretically, the addition of this KL penalty has been shown to be similar to performing Bayesian inference~\cite{Korbak2022RLWK} on the model.
A hyperparameter $\beta$ controls the weight of this KL penalty regularization during training. Further, it is common to compare different variants of RL algorithms at a fixed KL distance from the initial model, with the aim of maximizing the reward with the lowest possible KL divergence.

\subsubsection{Iterated RLHF}

As training progresses, the reward model can become miscalibrated with human preferences at higher rewards~\cite{HHHAssistant-Bai2022TrainingAH}.
This is because the reward model was trained on outputs from the initial model, which inherently have low-valued rewards. Consequently, several methods~\cite{SummarizeHF-Stiennon2020LearningTS,HHHAssistant-Bai2022TrainingAH} have employed an iterative training approach, where new outputs are generated by the updated policy, which are then annotated by humans for feedback. The reward model is then retrained based on this new human feedback, followed by training of the policy model.
This process, referred to as Iterated-RLHF or Online-RLHF, is repeated for several iterations. Although effective, this procedure is naturally expensive and time-consuming.

\paragraph{Limitations in Current RLHF practices}

Despite the impressive results achieved by RLHF in practice, it is an unstable training process~\cite{Choshen2019OnTW}. Moreover, it is highly sensitive to hyperparameters, necessitating a significant amount of hyperparameter tuning~\cite{rafailov2023direct, Yuan2023RRHFRR}. Furthermore, the generalization capabilities of RLHF and other issues, such as underperformance on metrics not captured by the reward model, warrant further investigation. A comprehensive examination of these aspects is discussed in Section~\ref{sec:rlhf_properties}.

%% file: LaTeX/survey_sections/260_properties_of_rlhf_models.tex
\subsection{Limitations of RLHF Models}
\label{sec:rlhf_properties}

Fine-tuning models using Reinforcement Learning from Human Feedback (RLHF) showcase a remarkable ability to align with human preferences and generalize to new scenarios and is more sample-efficient than supervised fine-tuning. Nonetheless, these models exhibit characteristics and behaviors that warrant careful consideration, prompting the need for further exploration and refinement.

\paragraph{Alignment Capabilities}
One intriguing property, referred to as the Alignment Tax, was identified by \cite{InstructGPT-Ouyang2022TrainingLM}. The phenomenon reveals that RLHF-trained chat models sometimes perform poorly compared to initial policy in downstream tasks, suggesting a cost linked to aligning human preferences.
To mitigate this, they propose incorporating the pre-training objective into RLHF-finetuning, which substantially reduces the Alignment Tax.
Moreover, \cite{HHHAssistant-Bai2022TrainingAH} indicates that larger models tend to exhibit lower alignment tax.
\cite{HHHAssistant-Bai2022TrainingAH} also observed that RLHF models better align with human preferences as the scales of both the reward model and policy model increase. It is noteworthy, however, that a similar scaling effect could be seen in instruction-finetuned SFT models. A comprehensive comparison of the scaling effects on RLHF versus SFT models is currently lacking in the literature and would make for an intriguing future study.

\paragraph{Generalization Capabilities}

RLHF models have exhibited impressive generalization capabilities beyond their training data, including generalization on new prompts and human feedback.
For instance, \cite{InstructGPT-Ouyang2022TrainingLM} demonstrates RLHF-tuned models answering coding questions and following instructions in multiple languages despite being finetuned only in English and with limited code-related prompts.
This suggests that the majority of a language model's capabilities are acquired during pre-training, and RLHF merely aligns these capabilities to elicit desired behavior.
However, this generalization can be a double-edged sword, potentially leading to undesirable outcomes, especially when the feedback signal is sparse. For instance, the initial LLaMA2 Chat Model\footnote{\url{https://huggingface.co/meta-llama/Llama-2-70b-chat-hf}}, when prompted "How to kill a process?" refused to answer, drawing ethical concerns, though the intended answer was about terminating a computer process. This behavior likely stems from the model's extended generalization from examples that trained it to reject violent queries. The example further highlights the problems of imperfect rewards leading to misgeneralization, as discussed in Section~\ref{sec:misalignment}.
Further, a distributional shift between prompts used for reward model finetuning and RLHF training can result in the policy model misaligning with human preferences~\cite{HHHAssistant-Bai2022TrainingAH}.
Further, during RL training, outputs are sampled from the language model, which is evaluated using the reward model. However, deviations in parameters used for sampling outputs from the model during inference from those in training can yield poor results~\cite{Ramamurthy2022IsRL}.
\future{Can talk about bias and red-teaming}

\paragraph{Diversity \& Biases of RLHF model outputs}

Another characteristic of RLHF models is their low entropy in output distribution~\cite{HHHAssistant-Bai2022TrainingAH}, which challenges generating diverse responses~\cite{kirk2023understanding}. This holds true for both seen and unseen datasets. To address this, entropy regularization techniques are introduced~\cite{WayOff-Jaques2019WayOB, DialogueHITL-Li2016DialogueLW} to amplify diversity in the action space, albeit not always resolving the issue~\cite{Raichuk2021WhatMF}.
While not conclusive, \cite{HHHAssistant-Bai2022TrainingAH} found that while RLHF models exhibit better sentiment towards all classes, they display similar biases to underlying LLMs when sampling with temperature < 1 (i.e., with low diversity samples). This could be attributed to their lower entropy. Furthermore, while pre-trained models often generate probabilities that are well-calibrated, RLHF models may lose this calibration. For instance, \cite{OpenAI2023GPT4TR} found that for pre-trained GPT-4, the probability of generating an answer is often directly proportional to the probability of it being correct. However, in the case of RLHF models, the distribution is skewed towards more likely answers.

\paragraph{Objective Misalignment}

While RLHF aims to align language models with human preferences and intentions, reward model misalignment is frequently possible. For instance, ~\citet{singhal2023long} finds that reward models provide higher rewards to longer outputs. Further,

\paragraph{Robustness and Safety}

It is imperative to note that the reward model is merely an imperfect proxy for real human preferences/feedback. Due to the lack of calibration and robustness of reward models~\cite{HHHAssistant-Bai2022TrainingAH}, over-optimizing against the reward model can render it an ineffective measure (Goodhart's Law). This phenomenon, known as Reward Overoptimization, has been studied in the context of language models by \cite{RewardOveroptimization-Gao2022ScalingLF, coste2023reward}.

Further, training RLHF models in practice is very difficult for practitioners owing to unstable training~\cite{Choshen2019OnTW}, hyperparameter sensitivity~\cite{Yuan2023RRHFRR, rafailov2023direct}, loading multiple models leading to high memory usage~\cite{santacroce2023efficient}. As a result, there have been significant efforts to simplify the training process by learning directly from the available feedback using simpler supervised finetuning objectives, as we discuss in Section~\ref{sec:alternatives}.

In conclusion, while RLHF substantially enhances the performance of LLMs and aligns them with human preferences, it is not without its limitations. These include, but are not limited to, issues such as text hallucination~\cite{McKenna2023SourcesOH}, bias and toxicity~\cite{Deshpande2023ToxicityIC, Ferrara2023ShouldCB, gupta2023bias}, and the generation of harmful text when probed~\cite{RedTeaming-Perez2022RedTL, Wei2023JailbrokenHD}. Despite significant improvements, these models are not fully aligned with human preferences, underscoring the need for continued research and development in this field.

%% file: LaTeX/survey_sections/350_sparse_rewards_in_rl.tex
\subsection{Enriching Reward Signals in Reinforcement Learning}

\paragraph{Challenges with Sparse Rewards in Traditional RL}

Reinforcement learning (RL) has conventionally employed delayed and sparse rewards, where agents receive scalar feedback at the end of a trajectory or episode \cite{Sutton2005ReinforcementLA}. While this approach is straightforward to implement and aligns with the task objective, it is not without its drawbacks. Sparse rewards can lead to sample-inefficient learning due to extensive exploration requirements \cite{Bellemare2016UnifyingCE}. Additionally, they may result in reward hacking, where agents exploit unintended strategies to maximize rewards without solving the intended task \cite{Ibarz2018RewardLF}. Underspecified rewards, which do not fully capture the desired behavior, can also yield suboptimal or degenerate solutions \cite{HadfieldMenell2017InverseRD}.

\paragraph{Enriching Reward Signals}

To mitigate the limitations of sparse rewards, researchers have explored various methods for providing richer feedback in environments with inherently sparse rewards. These approaches include reward shaping, where the original reward signal is augmented with additional feedback \cite{Ng1999PolicyIU, Grzes2017RewardSI}; intrinsic motivation, which encourages exploration and learning through internal rewards based on novelty, curiosity, or learning progress \cite{Oudeyer2007IntrinsicMS, Bellemare2016UnifyingCE, Pathak2017CuriosityDrivenEB}; and multi-objective optimization with multiple reward signals \cite{Roijers2013ASO, Roijers2016MultiobjectiveDP}. Hierarchical RL, which decomposes complex tasks into simpler subtasks with their own reward structures, has also been investigated \cite{Dietterich1999HierarchicalRL, Barto2003RecentAI}. Moreover, richer forms of feedback, such as learning from corrections \cite{Jain2015LearningPF, Bajcsy2017LearningRO}, demonstrations \cite{Rengarajan2022ReinforcementLW}, and language feedback \cite{Matuszek2012AJM, Fried2017UnifiedPM}, have proven beneficial.

\paragraph{Implications for RLHF in LLMs}

Current RLHF pipelines for LLMs primarily rely on sparse rewards provided at the end of an episode, with reward models trained using sparse preference-based feedback. Similar challenges observed in traditional RL have also been identified in RLHF-tuned LLMs. Some progress has been made in learning from feedback for multi-objective optimization \cite{RewardSoup-Ram2023RewardedST}, language feedback \cite{Scheurer2022TrainingLM}, corrective feedback \cite{Madaan2023SelfRefineIR, Shinn2023ReflexionAA}, and denser rewards \cite{FineGrainedRLHF-Wu2023FineGrainedHF}. Future research should explore the integration of these techniques to address the unique challenges in training LLMs with RLHF, potentially improving generalization and robustness.

%% file: LaTeX/survey_sections/400_reward_alternatives.tex
\subsection{Moving Beyond RL Training}
\label{sec:alternatives}

While RLHF has been very successful, it still results in unstable training~\cite{Choshen2019OnTW}, is hyperparameter sensitive~\cite{Yuan2023RRHFRR, rafailov2023direct}, has high memory usage~\cite{santacroce2023efficient} making it difficult for practitioners to actually use it. As a result, there have been significant efforts to simplify the training process by learning directly from the available feedback using simpler supervised finetuning objectives.

\subsubsection{Alternatives to RL using Reward Model}

Once, a reward model is trained, it is not necessary to perform the RLHF-based training. Instead, an alternate approach during inference is to sample multiple outputs from the LLM and rank them using the reward model~\cite{WebGPT-Nakano2021WebGPTBQ, Cobbe2021TrainingVT}. This is also called best-on-n sampling or rejection sampling. If sampling multiple outputs, it is important to ensure diversity of outputs by adjusting the sampling parameters (such as higher temperature). This approach is often considered as either a baseline or augmented with RLHF-trained models for better inference-time results.

Further, various works~\cite{RAFT-Dong2023RAFTRR, Yuan2023RRHFRR, song2023preference} use the trained reward model to rank multiple responses and use the signal from the ranked responses to train the policy model, without using an elaborate RL algorithm. In another line of work, RAD~\cite{deng2023rewardaugmented} uses weighted-decoding of tokens at inference, based on a separately trained reward model.

\subsubsection{Alternatives to RL without Explicit Reward Models}

In this section, we discuss alternative methods to align language models with human feedback that do not rely on reward models.
While RLHF-PPO has shown promising results, it suffers from sensitivity to hyperparameters, the need for training additional models, and potential misalignment of the reward model~\cite{FineGrainedRLHF-Wu2023FineGrainedHF,rafailov2023direct,Pang2022RewardGI, Zhu2023FineTuningLM, singhal2023long}. To address these issues, recent research has explored various techniques that directly incorporate human feedback into the training process, without relying on additional reward models.

A straightforward approach is supervised fine-tuning on positive demonstrations from human feedback, such as instruction-finetuned models~\cite{InstructGPT-Ouyang2022TrainingLM, vicuna2023, zhou2023lima}. However, this method does not utilize negative feedback, which is crucial for training robust models that can handle adversarial situations and identify and correct errors.

Recent works, such as \cite{Liu2023ChainOH, Zhang2023TheWO}, provide both positive and negative demonstrations/feedback and maximize the likelihood of generating positive/preferred output. These methods have shown better performance than RLHF methods on summarization and dialogue tasks. \citet{SLiC-HF-Zhao2023SLiCHFSL} demonstrate that Sequence Likelihood calibration (SLiC)~\cite{Zhao2022CalibratingSL} can be used to train models on off-policy offline data collected for different models, resulting in better performance than RLHF-based methods on summarization tasks. SLiC uses a ranking calibration loss that contrasts positive and negative sequences while motivating the model to predict the positive class. Further, RSO~\cite{liu2023statistical} improves policy learning in SLiC by using statistical rejection sampling from the policy.

\citet{rafailov2023direct, azar2023general} further reformulate the objective encoded in the RLHF PPO algorithm, and train the model directly on the new objective, without the need for a separate reward model. This follows the intuition, that the policy model can be implicitly used as a reward model for training itself based on the collected feedback. However, the results are preliminary, and extending to out-of-distribution prompts may not be possible without the introduction of an explicit reward model.

Another line of research focuses on refining model-generated responses using human-encoded principles or feedback. \cite{Bai2022ConstitutionalAH, kundu2023specific} propose a framework where a list of human-encoded principles (Constitution) guide the model to critique its generations and self-refine the responses. The model is then fine-tuned on the refined responses. Self-Align \cite{Sun2023PrincipleDrivenSO} follows a similar procedure but further removes the need to start with an RLHF-finetuned model. They fine-tune the pretrained LLaMA~\cite{Touvron2023LLaMAOA} base model using less than 300 lines of human feedback (in the form of constitutional principles) and achieve performance comparable to state-of-the-art models in terms of helpfulness and harmlessness.

Another direction of work learns to generate or select good feedback for model outputs and apply it to refine language model outputs.
\cite{Scheurer2022TrainingLM} takes a similar refinement approach but utilizes available summarization feedback. The initial model is conditioned on input, feedback, and output, generating multiple refinements. The model is then fine-tuned on refinements with the highest similarity to human feedback.
\cite{SecondThoughts-Liu2023SecondTA} aligns human moral values by modeling DP (dynamic-programming) based edits from unaligned source text to target aligned text. The model is then fine-tuned on the refinements generated by the edits, using RL for the second part of the process. \cite{Xu2022LearningNS} fine-tune a dialogue model using multi-modal feedback with the DIRECTOR method~\cite{Arora2022DirectorGF}, which models both negative and positive sequence labeling directly in the language model head.

In summary, these alternative methods generate new data based on feedback or guidelines and then use it to fine-tune the model. These approaches reduce the reliance on reward models and have shown promising results in some tasks, making them a viable alternative to RLHF-PPO. While these models are easier to train and help in alleviating many drawbacks of RLHF, the evaluation performed has been performed only on specific domains, and constrained settings. Moreover, other in-depth analysis such as sample efficiency and properties exhibited by these models, especially on out-of-distribution data needs to be explored further.

%% file: LaTeX/conclusion.tex
\section{Discussion and Conclusion}

In this work, we explore the fundamental aspects of reinforcement learning from human feedback (RLHF), aiming to clarify its mechanisms and limitations.
We highlight the underlying assumptions necessary for RLHF and examine the impact of different implementation choices, shedding light on the workings of this approach.
Our analysis naturally focuses on the reward models, which constitute the core component of RLHF. We introduce the concept of \textit{oracular rewards}, which represent the ideal reward signals that reward models should approximate.
The challenges encountered in learning these reward functions highlight both the practical and fundamental limitations of RLHF, as thoroughly analyzed by \citet{casper2023open}.

Our comprehensive review of the existing literature traces the development of RLHF from its inception to the recent advancements.
We cover various aspects: the types of feedback, the details and variations of training algorithms, and alternative methods for achieving alignment without using reinforcement learning.
In related work, \citet{kaufmann2023survey} extensively surveys RLHF, highlighting its evolution from preference-based learning.

Despite the numerous variations of RLHF, the core principle of learning from evaluative feedback remains unchanged.
This form of learning is naturally suited to reinforcement learning, while the specifics of agent formulation, the nature of reward feedback, and environment definition continue to evolve.
We anticipate the reduction of reliance on human (or AI) feedback by using existing knowledge sources to construct rewards, which is one of the most promising directions for future efforts to enhance the impact of RLHF.
Additionally, improving reward encoding mechanisms to better reflect the diversity of human preferences is an important area for further research.

As RLHF continues to advance and reach its full potential, supported by research in these areas, the use of LLMs is also expanding.
Until we fully understand the implications of RLHF, it is crucial to develop robust methods for quantifying uncertainty in the outputs generated by an LLM.
Such techniques would enable us to identify and address low confidence outputs, which is especially important in safety-critical applications.
Ultimately, understanding its implications becomes paramount as advancements in RLHF increasingly influence industries and economies.
Thus, research in this field is critically important in shaping the future of large-scale language modeling and its societal impact.

\paragraph{Acknowledgements} We thank Khanh Nguyen for extensive and insightful feedback on earlier versions of the draft. We also thank Wenlong Zhao, Tuhina Tripathi, and Abhiman Neelakanteswara for their help with improving the clarity of the manuscript.